\documentclass{article}

\usepackage[preprint]{neurips_2026}
\usepackage[utf8]{inputenc}
\usepackage[T1]{fontenc}
\usepackage{float}
\usepackage{amsmath,amssymb,amsthm,mathtools}
\usepackage{amsfonts}
\usepackage{bm}

\usepackage{booktabs}
\usepackage{tabularx}
\usepackage{enumitem}
\usepackage{algorithm}
\usepackage[noend]{algpseudocode}
\usepackage{graphicx}
\usepackage{subcaption}

\usepackage{nicefrac}
\usepackage{microtype}
\usepackage{xcolor}
\usepackage{url}
\usepackage{etoolbox}
\definecolor{darkgreen}{RGB}{0,100,0}
\usepackage{wrapfig}
\usepackage{hyperref}
\hypersetup{
    colorlinks=true,
    citecolor=blue,
    linkcolor=black,
    urlcolor=darkgreen
}

\usepackage[nameinlink,capitalize,noabbrev]{cleveref}

\newif\ifcomments
\commentsfalse   

\DeclareMathOperator*{\argmax}{arg\,max}

\newtheorem{assumption}{Assumption}[section]
\newtheorem{lemma}[assumption]{Lemma}

\newtheorem{theorem}[assumption]{Theorem}

\theoremstyle{remark}

\title{A Fused Gaussian Process Bandit Framework for Multi-Fidelity Bayesian Optimization}

\author{
Muyun Lu\textsuperscript{1}
\And
Jose E Aguilar Escamilla\textsuperscript{2}
\And
Huazheng Wang\textsuperscript{2}
\And
Ying Lin\textsuperscript{1}\thanks{Corresponding author: ylin53@Central.UH.EDU}
\\
\textsuperscript{1} Department of Industrial and Systems Engineering, University of Houston \\
\textsuperscript{2} School of  Electrical Engineering and Computer Science, Oregon State University
}
\date{}

\begin{document}

\maketitle

\begin{abstract}
Bayesian optimization (BO) is widely used to maximize expensive black-box functions when direct evaluation is costly. In many real-world applications, the high-fidelity objective can be approximated by a series of lower-fidelity models that provide partial but informative signals at substantially reduced cost. Multi-fidelity BO (MFBO) algorithms aim to strategically combine these information sources to achieve an optimal trade-off between accuracy and evaluation cost. However, existing MFBO methods often oversimplify the complex dependencies between fidelity levels or rely on heuristic acquisition functions that lack theoretical performance guarantees. To address these limitations, we introduce a new formulation of the MFBO problem as a fused Gaussian Process bandit and develop a corresponding Fused Gaussian Process Upper Confidence Bound (FGP-UCB) algorithm. Our approach accurately predicts the high-fidelity function by fusing multi-fidelity observations through an autoregressive Gaussian Process model and explicitly quantifies the uncertainty arising from parameter estimation across fidelity levels. The proposed method also provides theoretical simple-regret guarantees, overcoming a key limitation of existing MFBO algorithms that lack provable performance bounds. Effectiveness of the proposed method is demonstrated via both theoretical analysis and empirical studies.
\end{abstract}

\section{Introduction}\label{sec:Intro}

In many areas of computational science and engineering, optimizing expensive-to-evaluate black-box functions presents a critical challenge. For example, tuning the architecture and hyperparameters of machine learning models~\citep{snoek2012practical}, designing materials with desired physical properties~\citep{ueno2016combo,zhang2020bayesian}, and optimizing aerodynamic shapes in engineering design~\citep{marchildon2024gradient} all require repeated evaluations of costly black-box objectives. These complex systems often lack explicit mathematical forms or derivative information, rendering conventional gradient-based methods inapplicable. Furthermore, because individual function evaluations are computationally expensive, methods requiring numerous iterations are often impractical. To address these limitations, Bayesian optimization (BO) has emerged as a framework to efficiently optimize such functions with a minimal evaluation budget.

BO operates sequentially, employing a surrogate model and an acquisition function to determine the next evaluation point. The surrogate model, commonly a Gaussian Process (GP)~\citep{GP}, provides a probabilistic estimate of the high-fidelity function, capturing both predicted values and associated uncertainty. The acquisition function leverages this uncertainty to balance exploration (searching new regions) and exploitation (refining known promising areas). Popular acquisition strategies include Expected Improvement (EI)~\citep{EI}, Probability of Improvement (PI)~\citep{PI}, and the Upper Confidence Bound (UCB)~\citep{UCB1,UCB2,UCB3}. Among these, UCB-based methods provide a computationally explicit closed-form expression with a tunable parameter, granting direct control over the exploration-exploitation trade-off while offering strong theoretical guarantees regarding cumulative regret convergence.

Despite its data efficiency, standard BO may still require a substantial number of evaluations to converge. This poses significant challenges in domains such as advanced manufacturing or aerospace engineering, where querying the true high-fidelity function is prohibitively expensive. For instance, in the sequential experiment design of manufacturing processes, quality assessments often rely on offline experiments that are accurate but slow~\citep{zhang2023bayesianoptimisationsequentialexperimental}. Similarly, high-fidelity simulations, such as Computational Fluid Dynamics (CFD) models in aircraft design, are computationally intensive, limiting the applicability of traditional BO~\citep{Mackman}.

To mitigate these costs, researchers have proposed leveraging \textit{multi-fidelity} data by incorporating both high-fidelity (expensive, accurate) and low-fidelity (cheap, approximate) evaluations~\citep{MFMAB}. Sources such as in-line sensors or simplified simulations can provide faster, albeit less accurate, estimates of system quality. Multi-fidelity Bayesian Optimization (MFBO) exploits these heterogeneous data sources to accelerate convergence without sacrificing accuracy. The core strategy involves utilizing low-fidelity evaluations to broadly explore the design space, while reserving high-fidelity queries to refine solutions in promising regions~\citep{Mackman}.

Existing MFBO methodologies can be broadly categorized by their theoretical formulation: those rooted in the multi-armed bandit framework and those adopting decision-theoretic or surrogate-centric frameworks. Decision-theoretic and surrogate-centric approaches—including improvement-based and information-theoretic methods—prioritize selecting samples that maximize utility relative to cost or reduce epistemic uncertainty about the optimum. For example, Augmented Expected Improvement (AEI)~\citep{huang2006global, huang2006sequential} and extensions like Multi-Fidelity Knowledge Gradient (MF-KG)~\citep{poloczek2017multi} aim to maximize expected improvement per unit cost. Similarly, information-theoretic algorithms such as Multi-Fidelity Max-value Entropy Search (MF-MES)~\citep{takeno2020multi} emphasize querying points that offer the largest expected reduction in uncertainty regarding the global optimum. While often effective in practice, these methods do not explicitly adopt an optimism-in-the-face-of-uncertainty principle to guide the exploration–exploitation tradeoff. Instead, they prioritize expected utility or information gain, which can make it more difficult to systematically quantify exploration and establish theoretical performance guarantees. Moreover, computational challenges—such as the intractability of exact mutual information computation in MF-MES—or high data requirements in deep surrogate models (e.g., DNN-MFBO~\citep{DNN-MFBO, MFBNN}) limit their robustness in data-scarce or high-cost regimes.

To address the absence of an explicit optimism-driven exploration–exploitation mechanism in these frameworks, a separate line of work formulates MFBO as a multi-armed bandit problem. A representative method, MF-GP-UCB~\citep{MFGPUCB}, extends the classical GP-UCB algorithm by modeling each fidelity with an independent Gaussian Process and selecting queries via a cost-weighted upper confidence bound. This formulation provides sub-linear regret guarantees while allocating inexpensive low-fidelity evaluations for broad exploration prior to high-fidelity verification. However, MF-GP-UCB suffers from several limitations. First, its modeling framework treats fidelities independently and therefore fails to fully exploit the rich cross-fidelity correlations that commonly arise in practical applications. Second, its regret analysis relies on a uniformly bounded discrepancy assumption between fidelities, which can be restrictive in complex real-world systems where approximation errors vary substantially across the design space. Consequently, the resulting confidence bounds may become overly conservative, limiting the algorithm's ability to effectively leverage informative low-fidelity observations and reducing overall sample efficiency.

These limitations motivate the need for a new class of algorithms that combine the theoretical rigor of bandit-based regret minimization with the modeling strength of correlated multi-fidelity surrogates, such as the Autoregressive Gaussian Process (ARGP) or co-kriging models~\citep{cokriging}. While ARGP models have shown success in sparse modeling, leveraging them for multi-fidelity bandit optimization remains underdeveloped. A primary challenge is the structural dependency of the model: predicting high-fidelity responses at a specific location typically requires observed low-fidelity data; if this data is missing, substituted predictions introduce additional error. Furthermore, ARGP models are typically estimated via Expectation-Maximization (EM) or sampling algorithms \citep{cheng2021hierarchical, baillie2025multi}. Uncertainty of estimators from these algorithms are theoretically difficult to quantify, and misestimated uncertainty can severely undermine acquisition functions (such as UCB), leading to suboptimal fidelity and query selection.

To fill this gap, this paper proposes \textbf{Fused Gaussian Process Upper Confidence Bound (FGP-UCB)}, a novel MFBO algorithm within the bandit framework that addresses these limitations. Our method extends the autoregressive GP framework to a \textit{fused GP surrogate model} that jointly models high- and low-fidelity functions through their latent correlation structure and integrates it with the upper confidence bound (UCB) based acquisition strategy which provides sub-linear regret guarantees. FGP-UCB explicitly captures both the inter-fidelity correlation and residuals while accounting for parameter uncertainty in its predictions. Moreover, it formulates the MFBO problem under this fused GP model and introduces a new acquisition strategy grounded in the upper confidence bound (UCB) principle. A novel parameter, derived from posterior uncertainty and inspired by LinUCB-based regret analysis~\citep{abbasi2011improved,shallowprice}, is used to define the confidence bound. Based on this bound, FGP-UCB selects evaluation locations and determines fidelity levels using posterior variance. The algorithm strategically allocates high-fidelity evaluations to regions with high potential for improvement while leveraging low-fidelity observations to enable efficient exploration. We provide both theoretical and empirical justifications for our approach. Theoretically, we derive regret bounds showing that FGP-UCB requires fewer high-fidelity queries and converges more efficiently than single-fidelity or loosely coupled multi-fidelity strategies. Empirically, we demonstrate the performance of FGP-UCB on synthetic benchmarks and real-world applications. Our method consistently outperforms existing approaches in terms of sample efficiency and simple regret, especially in data-limited settings.

The remainder of this paper is structured as follows. Section 2 reviews the relevant literature, and Section 3 details the proposed FGP-UCB method. Section 4 presents empirical studies demonstrating the method's effectiveness, followed by Section 5, which concludes the work with final remarks and directions for future research.

\section{Literature Survey}\label{sec:Survey}

\subsection{Multi-Fidelity Bayesian Optimization}
Bayesian Optimization (BO) is a sequential design strategy for the global optimization of black-box functions that are expensive to evaluate, non-convex, and often noisy. Unlike gradient-based methods, BO does not require derivative information. Instead, it builds a probabilistic surrogate model of the objective function and employs an acquisition function to determine the most informative next point to evaluate, balancing the trade-off between exploration (sampling high-uncertainty regions) and exploitation (sampling regions predicted to be optimal)~\citep{garrido2025information}.

While standard BO is highly effective for optimizing expensive black-box functions, it remains constrained by the prohibitive cost of obtaining high-quality data points. In many practical engineering and scientific applications, sole reliance on these expensive high-fidelity evaluations—such as physical experiments or high-resolution simulations—can quickly exhaust the computational budget before a global optimum is found. Multi-Fidelity Bayesian Optimization (MFBO) addresses this limitation by introducing a cost-efficient hierarchy of information sources. By leveraging abundant, inexpensive low-fidelity approximations to explore the search space broadly and reserving expensive high-fidelity evaluations to correct biases and refine the search around promising regions, MFBO significantly accelerates convergence while minimizing total experimentation costs~\citep{alexandrov1998trust,huang2006sequential,cokriging,viana2014special}.

MFBO offers several distinct advantages for optimizing computationally expensive objective functions. By employing multi-fidelity surrogates, MFBO significantly reduces the requisite number of high-fidelity evaluations~\citep{fernandez2016review}. This is achieved by exploiting correlations within the mathematical structure or leveraging prior knowledge of the underlying physical processes. Furthermore, MFBO inherits the robust exploration-exploitation balance provided by standard BO acquisition functions, while the ability to parallelize low-fidelity evaluations further accelerates the optimization process~\citep{takeno2020multi}. Finally, the framework’s capacity for adaptive fidelity selection allows it to strategically determine when and where to sample high-fidelity data. This adaptability renders MFBO particularly effective in scenarios where the cost or availability of high-fidelity evaluations varies across the design space~\citep{poloczek2017multi}.

MFBO has been extensively applied to address complex optimization challenges across various scientific and engineering domains. In structural and manufacturing contexts, fidelity levels are often dictated by the mesh resolution in Finite Element Analysis (FEA). For example, \cite{huang2006sequential} utilized this approach for metal-forming process improvement, while \cite{tran2020smf} applied it to thermomechanical simulations for Flip-Chip package design.

The method is equally prominent in fluid dynamics and aerodynamic design. \cite{zhang2018variable} employed MFBO for airfoil optimization, distinguishing high- and low-fidelity models through differing convergence criteria and mesh resolutions within Reynolds-averaged Navier-Stokes (RANS) solvers. Similarly, \cite{serani2019adaptive} leveraged adaptive grid refinement to optimize NACA hydrofoils and minimize resistance in destroyer-type vessels. Other notable structural applications include the optimization of ship bases~\citep{shu2021multi} and aircraft panels~\citep{hao2020adaptive}.

\subsection{Multi-Fidelity Surrogate}
Multi-fidelity surrogate modeling is a strategic framework designed to accelerate engineering design and optimization by synthesizing data from sources of varying cost and accuracy. Instead of relying solely on computationally expensive high-fidelity simulations or potentially inaccurate low-fidelity approximations, this approach constructs a unified predictive model that exploits the correlation between them. By leveraging abundant, inexpensive data from low-fidelity models to establish global trends and using a sparse set of high-fidelity data to calibrate and correct local biases, multi-fidelity methods achieve high predictive accuracy with a significantly reduced computational budget compared to traditional single-fidelity approaches~\citep{wilke2024multifidelity}.

There are four typical classes of multi-fidelity surrogates to capture the correlations between fidelity levels: Independent Gaussian Processes, Composition, Input Augmentation, and Adjustment~\citep{do2025multifidelity}.

\subsubsection{Independent Gaussian Process.}
The Independent Gaussian Process approach is the simplest method, treating high-fidelity and low-fidelity models as entirely separate processes. This method constructs a standard GP model for the low-fidelity data and a separate GP model for the high-fidelity data, with no direct sharing of information or structure. While simple, it fails to capture the inherent correlation between fidelity levels; therefore, it is typically considered a baseline approach rather than a true multi-fidelity method.

\subsubsection{Composition.}
The class of composition methods comprises multi-fidelity techniques based on input-input and output-output mappings. Input-input mapping techniques~\citep{bandler1994space,bandler2004implicit,koziel2006space,robinson2008surrogate,tao2019input} typically take the form $f_{H}(x) = f_{L}(g(x))$, where $g(\cdot)$ maps the input space of the high-fidelity model to that of the low-fidelity model. Conversely, output-output mapping techniques~\citep{li2023nonlinearity,hebbal2021multi,cutajar2019deep,jiang2018space,zhou2017variable,zheng2014prior} generally take the form $f_{H}(x) = h(f_{L}(x))$, where $h(\cdot)$ maps the output space of the low-fidelity model to that of the high-fidelity model.

\subsubsection{Input augmentation.}
The Input Augmentation method approaches multi-fidelity optimization by incorporating the fidelity level as an additional input dimension within a unified surrogate model~\citep{perdikaris2017nonlinear,kandasamy2017multi,MFGPUCB}. Specifically, given a design input $x$, the augmented input is defined as $(x, m)$, where $m$ denotes the fidelity level (e.g., $m=0$ for low-fidelity and $m=1$ for high-fidelity). By treating all data sources within this joint space, the surrogate captures the correlation structure across fidelities. While structurally straightforward, this method faces challenges when the correlation between fidelity levels is weak or highly non-stationary.

\subsubsection{Adjustment.}
The class of adjustment-based multi-fidelity methods constructs surrogates that utilize adjustment factors to update low-fidelity models to represent high-fidelity data. These are commonly realized through Autoregressive or Hierarchical Gaussian Process (ARGP \& HGP) models~\citep{kennedy2000predicting}. This surrogate formulation can be described by $f_{H} = \rho(x)f_{L}(x) + f_{\delta}(x)$, where $\rho(x)$ is a link function serving as an adjustment coefficient, and $f_{\delta}(x)$ is a residual function quantifying the discrepancy between the adjusted low-fidelity model and the high-fidelity function.

Adjustment surrogates occupy a significant niche in multi-fidelity modeling due to two key advantages: ease of implementation and simplicity of analysis. The core concept is to model the residual error between the adjusted low-fidelity data and the high-fidelity data, making this approach highly compatible with established single-fidelity surrogate software packages.

Moreover, this structure provides a clear decomposition of the total error. The uncertainty in the final high-fidelity prediction is directly attributed to the uncertainties of the low-fidelity model, the adjustment coefficient, and the discrepancy model. This linearity facilitates the interpretation of each fidelity level's contribution and yields robust estimates of prediction variance, which is critical for sequential design and optimization algorithms. In this work, we employ adjustment-based methods as our multi-fidelity surrogate.

\subsection{Multi-Fidelity Acquisition Function}
A pivotal challenge in MFBO lies in formulating an effective acquisition function. In contrast to standard BO, which optimizes solely over the design space $x \in \mathcal{X}$, an MFBO acquisition function must navigate an augmented joint space comprising both the design parameters and fidelity levels, defined as the product space $(x, m) \in \mathcal{X} \times \mathcal{M}$. Consequently, the algorithm must simultaneously determine where to sample and at what fidelity to query the objective function, balancing information gain against computational cost. Generally, these acquisition strategies are classified into two primary categories: heuristic approaches and sequential selection methods.

\subsubsection{Heuristic approach}
Heuristic acquisition functions extend standard BO by operating within a joint space defined by the design parameters $x$ and the fidelity level $m$. These methods are typically derived by modifying classic "one-step look-ahead" acquisition functions—such as Expected Improvement (EI)—using auxiliary efficiency terms. These auxiliary terms explicitly balance the computational cost of a specific fidelity against its contribution to the predictive accuracy of the multi-fidelity surrogate. Due to their computational simplicity and ease of implementation, heuristic approaches remain a staple in engineering design optimization~\citep{huang2006sequential,ghoreishi2019multi,di2021multifidelity,grassi2023raal,poloczek2017multi}.

The evolution of these functions can be categorized by their underlying theoretical basis. Improvement-based methods were among the first to adapt classical strategies. A seminal example is the Augmented Expected Improvement (AEI), proposed by \cite{huang2006global, huang2006sequential}. AEI practically scales the standard EI value by the correlation between a lower fidelity and the high-fidelity ground truth, inversely weighted by the cost ratio.

Moving beyond myopic (one-step) heuristics, the Multi-Fidelity Knowledge Gradient (MF-KG), developed by \cite{poloczek2017multi}, offers a non-myopic perspective. Instead of looking only at immediate improvement, MF-KG calculates the expected increase in the maximum posterior mean of the high-fidelity function resulting from a prospective sample at $(x, m)$.

Currently, Information-Theoretic methods represent the most active research frontier. These approaches naturally quantify the "value of information" relative to its cost. A prominent method in this class is Multi-Fidelity Max-Value Entropy Search (MF-MES)~\citep{takeno2020multi}, which shifts the optimization goal from locating the design point $x$ to reducing uncertainty regarding the optimal function value $f^*$.

\subsubsection{Sequential selection} 
The sequential selection consists two steps. The first step is to select the new design point, then the second step is to select the fidelity with the design point found in the first step. The main difference between this approach and the heuristic approach is the selection of design point is independent of the selection of fidelity. This decoupling of selection of design point and fidelity is often motivated by the difficulty of optimizing mixed-integer spaces (continuous design point $x$ and discrete fidelity level $m$) and the desire to align the search strategy with the target (optimizing the high-fidelity function). Then the solution improvement is separated from the consideration of computational cost and the accuracy improvement of the multi-fidelity surrogate.

The most prominent literature utilizing this sequential structure stems from the generalization of the Multi-Armed Bandit problem to Gaussian Processes, specifically through the work of \cite{MFMAB,kandasamy2017multi,MFGPUCB}. The method in these works uses an Upper Confidence Bound strategy. The key insight is to define a "fidelity threshold." The algorithm only queries the high-fidelity source if the Upper Confidence Bound of the high-fidelity model is higher than the current best and the uncertainty at lower fidelity levels is sufficiently reduced.

\cite{chen2017multimodel} generated fictitious samples for fidelity $m$ at location $x$ by drawing from the Gaussian distribution characterized by the current predictive mean and variance. They quantified the reduction in standard deviation for fidelity $m$ as the difference between the fused model's current predictive standard deviation at $x$ and the expected posterior standard deviation following an update with the fictitious sample. \cite{tran2020smf,tran2020multi} employed a similar strategy for fidelity querying.

Our work prioritizes a sequential selection framework due to its significant advantages over heuristic approaches in terms of efficiency, robustness, and theoretical tractability.

Computationally, sequential methods reduce the search burden by optimizing solely over the design space $\mathcal{X}$, avoiding the expensive joint optimization over $\mathcal{X} \times \mathcal{M}$ required by heuristic methods.Furthermore, sequential strategies separate location selection from cost-dependent fidelity allocation, allowing high-fidelity evaluations to be directed toward regions of interest. Finally, this structure facilitates regret analysis by separating the optimization of the high-fidelity objective from the fidelity-selection decision~\citep{MFGPUCB}.

However, a critical gap remains in the field: relatively little work has investigated sequential selection algorithms specifically designed for adjustment-based surrogate models.

\section{Methodology}\label{sec:Method}

\subsection{MFBO Problem Statement}
Bayesian optimization (BO) is a sequential design strategy aimed at maximizing an unknown, expensive-to-evaluate black-box function $f_{H}:\mathcal{X}\rightarrow \mathbb{R}$, subject to a finite budget $\Lambda$. The search space $\mathcal{X} \subseteq [0,1]^{d}$ is typically a finite or compact domain. At each round $t$, the optimizer selects a query point $x\in \mathcal{X}$ to evaluate. Each query to the high-fidelity function $f_H$ incurs a cost $c_H$ and yields a noisy observation $y_{H} = f_{H}(x)+\epsilon_{H}$, where the noise term is sub-Gaussian random noise with zero-mean, e.g. $\epsilon_{H}\sim N(0,\sigma_{H}^{2}),\mathbb{E}[\epsilon_{H}]=0$.

The fundamental goal of BO is to identify the global maximizer $x^{*} = \text{argmax}_{x\in \mathcal{X}} f_{H}(x)$ before the total budget $\Lambda$ is exhausted. To achieve this, the algorithm must maintain a delicate balance between exploitation—querying regions where $f_H$ is expected to be high—and exploration—querying regions of high uncertainty to improve the model. Efficient algorithms must navigate this trade-off to ensure the optimum is located within the cost constraints.

However, in many practical applications, querying the high-fidelity function $f_H$ is prohibitively expensive, leading to sparse data and insufficient evaluations to guarantee convergence. To overcome this data scarcity, one can leverage low-fidelity approximations (e.g., simplified simulations or coarse-grained models) to guide the search. Let $f_{L}:\mathcal{X}\rightarrow \mathbb{R}$ denote a low-fidelity approximation of the target. When a point $x$ is queried on this lower fidelity, the user obtains an observation $y_{L} = f_{L}(x) + \epsilon_{L}$ ($\epsilon_{L}\sim N(0,\sigma_{L}^{2})$) at a cost $c_{L}$, such that $c_{L} \ll c_{H}$. While cheaper, these observations typically suffer from higher noise variance ($\sigma_{L} > \sigma_{H}$) or structural bias compared to the target.

This setting formalizes the Multi-fidelity Bayesian Optimization (MFBO) problem. At each iteration $t$, the policy must jointly select a candidate point $x_{t}$ and a fidelity level $m_t \in \{L,H\}$ to generate an observation $y_{m_t}(x_t)$. The core challenge in MFBO is effectively fusing information from cheap, low-fidelity queries to explore the space, while strategically reserving the budget for high-fidelity queries to accurately exploit and verify the optimum.

The performance of an MFBO algorithm is measured by its ability to minimize the simple regret $S(\Lambda)$ on the high-fidelity function $f_{H}$, which is a standard theoretical performance criterion in MFBO and stochastic bandit optimization \citep{bubeck2011pure, vakili2021optimal}. It measures the quality of the design recommended by the algorithm with respect to the underlying latent high-fidelity objective, rather than the noisy observations. This formulation is appropriate because the objective of Bayesian optimization is to identify the maximizer of the latent function, whereas the observed responses are stochastic realizations corrupted by measurement noise and therefore do not faithfully represent the true objective values. By evaluating performance with respect to the latent objective, simple regret quantifies the intrinsic optimization error of the recommendation, which remains random because the recommendation depends on noisy observations, providing a stable and theoretically meaningful measure of algorithmic performance. It is important to note that while low-fidelity queries assist in learning the structure of the objective function, they provide no direct optimization reward. Unlike the best-query simple regret considered in MF-GP-UCB~\citep{MFGPUCB}, which evaluates the best high-fidelity point queried by the algorithm, our definition evaluates the design explicitly recommended by the algorithm. Since the best queried point cannot generally be identified from noisy observations without knowledge of the latent function values, this recommendation-based formulation provides a more direct assessment of the quality of the solution returned in practice.

Let $x^* \in \arg\max_{x\in\mathcal X}f_H(x)$ denote a global maximizer, with $f^*=f_H(x^*)$. Given an evaluation budget $\Lambda$, let $T(\Lambda)$ denote the number of completed queries and $\mathcal I_H(\Lambda)=\{t\leq T(\Lambda):m_t=H\}$ denote the indices of high-fidelity queries. The algorithm finally recommends a design $x_{out}$ as the optimal solution. Provided $\mathcal I_H(\Lambda)\neq\varnothing$, simple regret is defined as
\begin{equation}
\begin{aligned}
\label{SR}
S(\Lambda) = f^{*}-f_H(x_{out})
\end{aligned}
\end{equation}

Since the objective function is assumed to be corrupted by stochastic observation noise, the high-fidelity function values are generally unavailable during optimization and the simple regret is therefore not directly observable in real-world applications. Although realized value of simple regret cannot be computed in practice, it characterizes the latent quality of the design identified from noisy observations and therefore directly reflects the reliability of the algorithm’s recommendation. A bound on simple regret provides theoretical assurance that the algorithm can identify a near-optimal design despite stochastic observation noise. Consequently, simple regret is widely adopted in the theoretical analysis of Bayesian optimization algorithms and is commonly evaluated on synthetic benchmark problems where the ground-truth objective is known \citep{MFGPUCB, hu2025adjusted}. For practical applications where the ground-truth objective is unavailable, optimization performance is instead assessed using observable quantities, such as the observed high-fidelity objective value at the recommended design obtained under the available evaluation budget \citep{hernandez2016general, hvarfner2024vanilla}. Throughout this paper, simple regret is used as the theoretical performance criterion, while the real-world study is evaluated using the observed high-fidelity objective value at the recommended design.

Despite the widespread use of MFBO in practice, theoretical guarantees on performance—especially in terms of simple regret—remain largely underdeveloped. Most existing MFBO algorithms are heuristics: they select fidelity levels based on empirical correlations or cost heuristics, but their simple regret is not theoretically bounded, meaning convergence to the true high-fidelity optimum cannot be guaranteed under finite budget~\citep{huang2006global, huang2006sequential,poloczek2017multi,takeno2020multi}. A few recent works provide simple-regret guarantees, but they rely on restrictive assumptions, such as known and fixed correlations across fidelity levels or stringent bounds on the residual error between fidelity levels~\citep{MFMAB,kandasamy2017multi,MFGPUCB}. These assumptions rarely hold in real-world applications where fidelity relationships are nonlinear, nonstationary, or sample-dependent. This underscores the need for new MFBO frameworks that can provide provable simple-regret guarantees without relying on unrealistic assumptions about cross-fidelity structure or deterministic residual bounds.

\subsection{Fused Gaussian Process Bandit Framework}
\begin{figure}
    \centering
    {\includegraphics[width=0.7\textwidth]{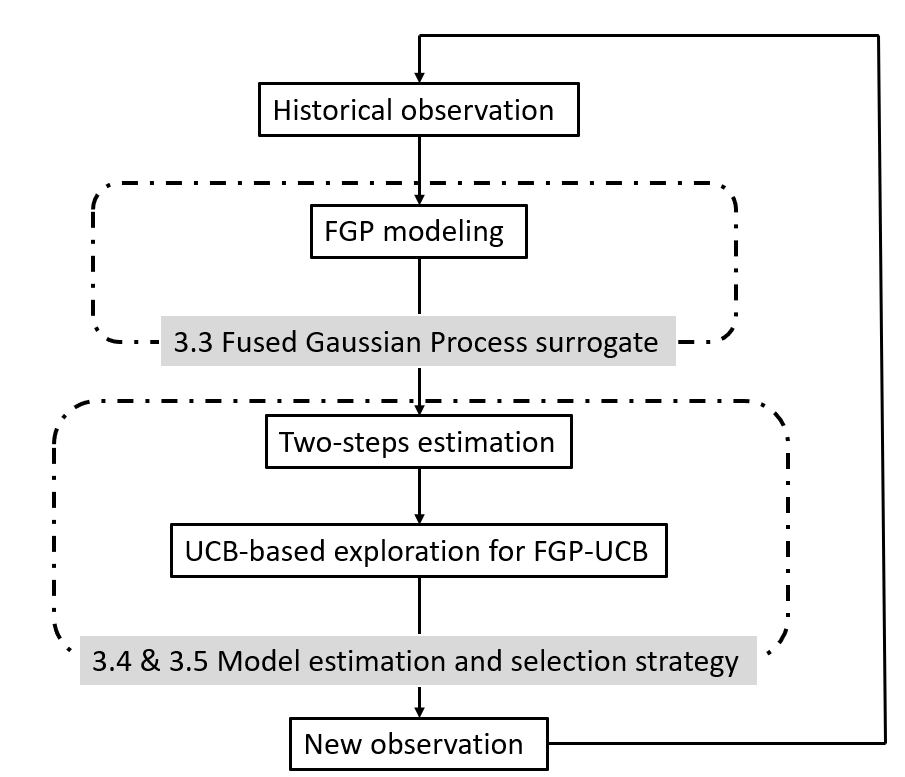}}
    \caption{The overall framework of the proposed Fused Gaussian Process Bandit}
    \label{framework}
\end{figure}

As illustrated in Figure \ref{framework}, the proposed Fused Gaussian Process Bandit framework comprises two primary components: Fused Gaussian Process surrogate modeling and a sequential decision-making strategy. In the modeling phase, we develop a novel adjustment-based multi-fidelity approach. Specifically, a two-step estimation process is employed to calibrate the FGP parameters, enabling the simultaneous quantification of the predictive mean and uncertainty across both low- and high-fidelity functions. Subsequently, an Upper Confidence Bound (UCB) acquisition function is designed to identify the optimal query configuration—sequentially selecting the input location and fidelity level—by synthesizing these predictive statistics. Finally, new observations are acquired and assimilated into the dataset, iteratively updating the model and refining the selection strategy for subsequent cycles.

\subsection{Fused Gaussian Process Surrogates}
To capture the relationship between fidelity levels while facilitating rigorous regret analysis, we assume that the high-fidelity objective function $f_{H}(x)$ is correlated with the low-fidelity function $f_{L}(x)$ via a fusion model:
\begin{equation}\label{fusedeq}
\begin{aligned}
f_{H}(x) = \rho(x) f_{L}(x) + f_\delta(x)
\end{aligned}
\end{equation}
$\rho(x) \in \mathbb{R}$ is an input-dependent link function characterizing the relationship between fidelity levels. The term $\rho(x)f_L(x)$ represents the adjusted low-fidelity component, and $f_\delta(x)$ is a residual function capturing the discrepancy between the adjusted low-fidelity component and the high-fidelity objective $f_H(x)$. We model the low-fidelity and residual functions using Gaussian Processes and parameterize the link function as follows:
\begin{enumerate}
\item \textit{Low-Fidelity Function}: The low-fidelity is modeled by a standard GP: $f_{L}(x) \sim \mathcal{GP}(0,k_{L}(x,x'))$, where the mean function is set to zero and the kernel $k_{L}(x,x')$ encodes the covariance structure.

\item \textit{Link Function}: The link function varies linearly with the input, $\rho(x)=\rho^{T}x$, where $\rho\in\mathbb{R}^{d}$ is an unknown parameter vector. This formulation provides a simple approximation when the scaling between fidelities varies approximately linearly over the design domain. Moreover, the linear structure enables closed-form parameter estimation and explicit quantification of estimation uncertainty~\cite{le2014recursive}.

\item \textit{Residual Function}: The residual function is modeled as an independent Gaussian Process, $f_{\delta}(x)\sim\mathcal{GP}(0,k_{\delta}(x,x'))$, to capture the discrepancies between the adjusted low-fidelity function and the high-fidelity objective.
\end{enumerate}

Following the recursive formulation of autoregressive multi-fidelity models \citep{le2014recursive}, we first estimate the low-fidelity posterior mean $\hat{\mu}_{L}(x)$ and use it to approximate the low-fidelity function values in the fusion model. Unlike standard Co-kriging \citep{cokriging}, which models the joint distribution of multi-fidelity observations together, this sequential formulation estimates the low- and high-fidelity models separately. Therefore, the fusion model can be represented as: 
\begin{equation}\label{fusedmueq}
\begin{aligned}
f_{H}(x) = \rho(x)\hat{\mu}_{L}(x) + f_{\delta}(x)
\end{aligned}
\end{equation}
This posterior-mean approximation accommodates non-collocated data and significantly reduces the computational overhead associated with large covariance matrices. Specifically, while standard Co-kriging scales cubically with the total number of observations—$\mathcal{O}((n_L+n_H)^3)$—due to the inversion of the joint covariance matrix, treating the fidelities sequentially reduces the computational complexity to $\mathcal{O}(n_L^3 + n_H^3)$.

Our primary contribution lies in leveraging this decoupled structure to derive rigorous confidence bounds for link-function estimation and high-fidelity prediction, which provide the foundation for subsequent regret analysis. By assuming a linear link function alongside GP models for both the low-fidelity and residual functions, we can explicitly isolate and quantify the uncertainty originating from multiple sources. Connecting this linear parameterization to established linear bandit concentration bounds \citep{abbasi2011improved} enables us to translate this multi-source uncertainty into a novel Upper Confidence Bound (UCB) score. Ultimately, integrating this new UCB score with our sequential fidelity-selection algorithm yields a provable theoretical regret guarantee for multi-fidelity Bayesian optimization.

\subsection{Model Inference}

To estimate the fused Gaussian Process surrogate using the low- and high-fidelity datasets, i.e. $\mathcal{D}_L=(X_L,Y_L)$ and $\mathcal{D}_H=(X_H,Y_H)$, we employ a sequential two-step estimation strategy. First, the low-fidelity GP is fitted using low-fidelity observations. Then, the link function and residual GP are estimated using the low-fidelity predictions and high-fidelity observations.

\noindent \textit{Step 1. Low-Fidelity Function Estimation}

The dataset $\mathcal{D}_L$ consists of noisy low-fidelity samples $Y_{L} = [y_{L,1},...,y_{L,n_{L}}]^{T}$ at input locations $X_{L} = [x_{L,1},...,x_{L,n_{L}}]^{T}$, where $n_{L}$ is the size of $\mathcal{D}_{L}$. The observations follow a standard GP model with Gaussian noise, i.e. $y_{L,i}=f_L(x_{L,i})+\epsilon_{L,i}$, $\epsilon_{L,i} \sim N(0,\sigma_L^2)$. Conditioning the GP prior on the dataset $\mathcal{D}_{L}$ yields a closed-form posterior distribution over $f_L(x)$, fully characterized by the predictive mean $\hat{\mu}_{L}(x)$ and predictive variance $\hat{\sigma}_{L}^{2}(x)$, given by:
\begin{equation}\label{muL}
\begin{aligned}
&f_{L}(x)|\mathcal{D}_{L} \sim N(\hat{\mu}_{L}(x),\hat{\sigma}_{L}^{2}(x))\\
&\hat{\mu}_{L}(x) = k_{L,n_{L}}(x)^{T}(K_{L}+\sigma_{L}^{2}I)^{-1}Y_{L}\\
&\hat{\sigma}_{L}^{2}(x)=k_{L}(x,x)-k_{L,n_{L}}(x)^{T}(K_{L}+\sigma_{L}^{2}I)^{-1}k_{L,n_{L}}(x)
\end{aligned}
\end{equation}
where $k_{L,n_{L}}(x) = [k_{L}(x_{L,1},x),...,k_{L}(x_{L,n_{L}},x)]^{T}$, and $K_{L}=[k_L(x_{L,i},x_{L,j})]_{i,j=1}^{n_L}$ is the low-fidelity kernel matrix.

\noindent \textit{Step 2. Link and Residual Functions Estimation}

Let the set of high-fidelity observations be $\mathcal{D}_{H} = (X_H, Y_H)$, where the input locations are $X_H = [x_{H,1}, \ldots, x_{H,n_H}]^T$ and the corresponding outputs are $Y_H = [y_H(x_{H,1}), \ldots, y_H(x_{H,n_H})]^T$.
Here, $n_H$ denotes the number of high-fidelity samples. The high-fidelity inputs $X_H$ do not need to coincide with the low-fidelity inputs $X_L$; the two datasets may be entirely non-overlapping.

Based on the fusion model assumption, we construct a plug-in high-fidelity GP surrogate $f_{H}(x)\sim \mathcal{GP}(\rho(x)\hat{\mu}_{L}(x), k_{\delta}(x,x'))$. The link function is characterized by an unknown parameter vector $\rho$ that needs to be estimated from data. Given the estimated low-fidelity posterior mean $\hat{\mu}_{L}$, the residual GP covariance matrix $K_{\delta}$, and the high-fidelity observations, the parameter vector can be estimated by maximizing the following regularized log-likelihood function:

\begin{equation}
\begin{aligned}
l_{\lambda}(\rho|\mathcal{D}_{H}, \hat{\mu}_{L}, K_{\delta}) = &-\frac{1}{2}\log|K_{\delta}+\sigma_{H}^{2}I| -\frac{\lambda}{2}\|\rho\|_{2}^{2}\\
&- \frac{1}{2}(Y_{H}-\rho(X_{H})\hat{\mu}_{L}(X_{H}))^{T}(K_{\delta}+\sigma_{H}^{2}I)^{-1}(Y_{H}-\rho(X_{H})\hat{\mu}_{L}(X_{H}))
\end{aligned}
\end{equation}

Under the assumption that the link function $\rho(x)$ follows a linear structure, a closed-form estimator for the parameter vector $\rho$ can be obtained. The resulting estimator, denoted as $\hat{\rho}$, together with the predictive mean $\hat{\mu}_{H}(x)$ and plug-in predictive variance $\hat{\sigma}_{H}^{2}(x)$ of the high-fidelity function, are given by:

\begin{equation}\label{eq3}
\begin{aligned}
&\hat{\rho} = (F_{L}^{T}(K_{\delta}+\sigma_{H}^{2}I)^{-1}F_{L} + \lambda I)^{-1}F_{L}^{T}(K_{\delta}+\sigma_{H}^{2}I)^{-1}Y_{H}\\
&\hat{\mu}_{H}(x) = (\hat{\rho}^{T}x)\hat{\mu}_{L}(x) + k_{\delta,n_{H}}(x)^{T}(K_{\delta}+\sigma_{H}^{2}I)^{-1}(Y_{H} - F_{L}\hat{\rho})\\
&\hat{\sigma}_{H}^{2}(x)=k_{\delta}(x,x)-k_{\delta,n_{H}}(x)^{T}(K_{\delta}+\sigma_{H}^{2}I)^{-1}k_{\delta,n_{H}}(x)
\end{aligned}
\end{equation}
Here, $\lambda>0$ is a regularization parameter ensuring that the matrix $F_{L}^{T}(K_{\delta}+\sigma_{H}^{2}I)^{-1}F_{L} + \lambda I$ is positive definite. The vector $k_{\delta,n_{H}}(x)$ is defined as $[k_{\delta}(x_{H,1},x),...,k_{\delta}(x_{H,n_{H}},x)]^{T}$, and $K_{\delta}=[k_{\delta}(x_{H,i},x_{H,j})]_{i,j=1}^{n_H}$ is the residual GP kernel matrix. The design matrix $F_{L}$ is constructed by multiplying each high-fidelity input vector by its corresponding low-fidelity posterior mean: \[F_{L} = \begin{pmatrix} \hat{\mu}_{L}(x_{H,1})*(x_{H,1})_{1} & ... & \hat{\mu}_{L}(x_{H,1})*(x_{H,1})_{d}\\ &... & \\\hat{\mu}_{L}(x_{H,n_{H}})*(x_{H,n_{H}})_{1} & ... & \hat{\mu}_{L}(x_{H,n_{H}})*(x_{H,n_{H}})_{d} \end{pmatrix}\]

The resulting high-fidelity surrogate in Eq.\eqref{eq3} combines the low-fidelity posterior mean with adjustments by estimated link function and the residual GP. It enables the high-fidelity model to capture the global trend learned from low-fidelity observations, while correcting for local bias through the residual process and input-dependent adjustments. The following section details the uncertainty quantification for each component of the high-fidelity prediction.

\subsection{Fused Gaussian Process-based Upper Confidence Bound}
In this section, we develop a novel Fused Gaussian Process-based Upper Confidence Bound (FGP-UCB) algorithm as the acquisition function to jointly optimize the selection of design points and fidelity levels for minimizing simple regret. The algorithm design begins with analyzing three sources of uncertainty in the high-fidelity prediction in Eq.~\eqref{eq3}: low-fidelity prediction uncertainty, link-function estimation uncertainty, and residual GP prediction uncertainty. The adjusted low-fidelity component plays a central role in high-fidelity prediction by transferring information from low-fidelity observations. Importantly, because the link parameters are estimated using low-fidelity posterior predictions, errors in these predictions can propagate into link function estimation, motivating a joint analysis of the two sources of error. In particular, we derive an estimation-error bound for the estimated adjusted low-fidelity component $(\hat{\rho}^{T}x)\,\hat{\mu}_{L}(x)$ and use this result to formulate a unified confidence bound for the high-fidelity prediction. This leads directly to the proposed FGP-UCB criterion. Building on this confidence bound, we design a sequential acquisition strategy that first selects the most promising design point~$x$ and then determines the most appropriate fidelity level to query, thereby achieving a principled balance between exploration and exploitation. While the current formulation focuses on a bi-fidelity setting, the proposed framework can be theoretically generalized to hierarchies with three or more fidelity levels via a chain of autoregressive models (e.g., modeling fidelity $m$ as a linear transformation of fidelity $m-1$). However, extending the regret guarantees to such a chain needs additional analysis, particularly regarding the recursive propagation of estimation errors through multiple link functions and the resulting compound uncertainty in the upper confidence bound.

\subsubsection{Incomplete Uncertainty Quantification}

Standard Upper Confidence Bound (UCB) criterion relies on two key components: a predictive mean and its associated epistemic uncertainty. While the predictive mean of high-fidelity response is defined in Eq.~\eqref{eq3}, quantifying its uncertainty in this multi-fidelity setting is considerably more complex. The posterior variance in Eq.~\eqref{eq3} cannot fully capture the high-fidelity prediction uncertainty, as the total uncertainty comprises three sources: (i) bias inherent in the low-fidelity prediction, (ii) uncertainty in estimating the link function, and (iii) uncertainty in residual function prediction. The uncertainty in the residual function prediction is directly captured by its posterior variance $\hat{\sigma}_{H}^{2}(x)$ in Eq.~\eqref{eq3}. However, deriving the uncertainty associated with the adjusted low-fidelity prediction $(\hat{\rho}^{T}x)\,\hat{\mu}_{L}(x)$ is nontrivial because it couples the variance of the low-fidelity estimator with the estimation error of the link function—which itself depends on the low-fidelity estimator.

\paragraph{Link Function Estimation Error}We address this challenge by isolating the effects of the predicted and ground-truth low-fidelity models within the estimated link parameter $\hat{\rho}$ in Eq.~\eqref{eq3}. This allows us to separate estimation uncertainty from prediction bias. The decoupled expression is then used to quantify the gap between $\hat{\rho}$ and the ground truth $\rho^{*}$, establishing a bound that incorporates both sources of error. The estimation bound for $\rho$ is concluded in Lemma \ref{lemma1}. This bound deviates from the standard result for linear contextual bandits established in \cite{abbasi2011improved} due to the introduction of a bias term.

\begin{lemma}[Link Function Estimation Bound]\label{lemma1}
Let $\|\rho^{*}\|_2\leq M_{\rho}$, $\lambda>0$, and
\(
A_L=F_L^TQ^{-1}F_L+\lambda I,
\)
where $Q=K_{\delta}+\sigma_H^2I$ is positive definite. With high probability $1-\delta$,
\begin{equation}
\begin{aligned}
\lVert \hat{\rho}-\rho^{*}-A_L^{-1}F_L^TQ^{-1}&(F_L^{*}-F_L)\rho^{*}\rVert_{A_{L}} \\ &\leq \sigma_{H}\sqrt{2\log\left(\frac{\det(A_{L})^{1/2}\det(\lambda I)^{-1/2}}{\delta}\right)} +\lambda^{1/2}M_{\rho} = \alpha_\rho
\end{aligned}
\end{equation}
Here, $F_L^{*}$ denotes the ground truth low-fidelity design matrix, $\rho^{*}$ represents the ground truth parameters of the link function
\end{lemma}

Proof of Lemma~\ref{lemma1} is provided in the Appendix. The term involving $F_L^*-F_L$ captures how using the low-fidelity posterior mean at the high-fidelity observation locations shifts the link-parameter estimate. Then the whole left-hand side measures the estimation error that remains after subtracting this shift. The right-hand side gives a high-probability radius $\alpha_\rho$ for this error. The radius combines a concentration term that accounts for stochastic estimation error which reflects the information in the data, and a bound for regularization bias.

At a design $x$, a change in $\rho$ affects the adjusted low-fidelity prediction through its inner product with $x\hat{\mu}_L(x)$. By Cauchy--Schwarz, the remaining error in estimating $\rho$ contributes at most $\alpha_\rho\lVert x\hat{\mu}_L(x)\rVert_{A_L^{-1}}$ to the adjusted low-fidelity prediction error at $x$. Theorem~\ref{theorem1} states the confidence bound for the full high-fidelity prediction.

\begin{theorem}\label{theorem1}
Let $\|\rho^*\|_2\leq M_\rho$ and let $A_{L,t}=F_{L,t}^TQ_t^{-1}F_{L,t}+\lambda I$, where $Q_t=K_{\delta,t}+\sigma_H^2I$, be the regularized information matrix. With high probability $1-\delta$,
\begin{equation}
\begin{aligned}
\lVert\hat{\mu}_{H,t}(x)-f_{H}(x)\rVert \leq \alpha_{\rho,t}\lVert x\hat{\mu}_{L,t}(x)\rVert_{A_{L,t}^{-1}} + \sqrt{\beta_{t}} \hat{\sigma}_{\delta,t}(x)
\end{aligned}
\end{equation}
where the scaling factor is defined as $\alpha_{\rho,t} = \lambda^{\frac{1}{2}}M_\rho + \sigma_{H}\sqrt{2\log(\frac{\det(A_{L,t})^{\frac{1}{2}}\det(\lambda I)^{-\frac{1}{2}}}{\delta})}$, and $\beta_{t}$ is the exploration parameter scaling the confidence interval width to ensure coverage probability.
\end{theorem}

Proof of Theorem \ref{theorem1} is provided in the Appendix. Theorem \ref{theorem1} demonstrates that, with high probability, the prediction error of the high-fidelity—derived from the proposed fused Gaussian Process surrogate—is bounded by a decomposition of uncertainty terms. Specifically, the bound consists of the estimation uncertainty of the adjusted low-fidelity function, represented by the weighted norm $\alpha_{\rho,t}\left\lVert x\hat{\mu}_{L,t}(x)\right\rVert_{A_{L,t}^{-1}}$ (where $A_{L,t}$ captures the accumulation of information in the link structure), and the estimation uncertainty of the residual function, denoted by $\sqrt{\beta_{t}} \hat{\sigma}_{\delta,t}(x)$. The detailed proof is provided in the Appendix.

\subsubsection{Fused Gaussian Process Acquisition Function}
\label{sec:FGPAQ}

Leveraging Theorem \ref{theorem1}, we propose a novel Fused Gaussian Process-based Upper Confidence Bound (FGP-UCB) criterion, defined in Eq.\eqref{UCB}. The FGP-UCB criterion ensures efficient budget allocation by balancing two competing objectives in the high fidelity. The mean term, $\hat{\mu}_{H,t}(x)$, exploits the potential optimal locations with high predicted responses in high-fidelity, while the uncertainty terms promote exploration of uncertain areas with high prediction variation. This criterion serves as the metric for evaluating the potential benefit of querying a specific location $x$.
\begin{equation}\label{UCB}
\begin{aligned}
\text{FGP-UCB}_t(x) = \hat{\mu}_{H,t}(x) +\alpha_{\rho,t}\lVert x\hat{\mu}_{L,t}(x)\rVert_{A_{L,t}^{-1}} + \sqrt{\beta_{t}} \hat{\sigma}_{\delta,t}(x)
\end{aligned}
\end{equation}

Let
\(
w_t(x)=
\alpha_{\rho,t}\|x\hat{\mu}_{L,t}(x)\|_{A_{L,t}^{-1}}
+\sqrt{\beta_t}\hat{\sigma}_{\delta,t}(x)
\)
denote the confidence radius in Eq.\eqref{UCB}. The corresponding Fused Gaussian Process-based Lower Confidence Bound (FGP-LCB) can be obtained as:
\begin{equation}\label{eq:fgp_lcb}
\mathrm{FGP\text{-}LCB}_t(x)
=
\hat{\mu}_{H,t}(x)-w_t(x).
\end{equation}
Both bounds are computed from observations available before iteration $t$. FGP-UCB is used for query selection, while FGP-LCB is used for the final recommendation. The UCB criterion balances exploration and exploitation by considering both the predicted high-fidelity response and its associated uncertainty, thereby guiding queries toward potentially optimal locations. In contrast, the LCB criterion provides a conservative estimate of the high-fidelity response, allowing the algorithm to recommend a queried design with a high predicted response and relatively low uncertainty.

In our multi-fidelity Bayesian optimization framework, the acquisition process is sequential: it first identifies the query location $x_t$ and subsequently selects which fidelity to query. As previously established, regrets are defined exclusively on the high-fidelity, while low-fidelity queries serve to elucidate the structure of the objective function. Consequently, the location that best balances the exploitation of maximal high-fidelity response and exploration of uncertainty in high-fidelity is queried. This is achieved by selecting the location with maximal FGP-UCB criterion:
\begin{equation}\label{opt_acq}
x_{t} = \operatorname{argmax}_{x} \text{FGP-UCB}_t(x)
\end{equation}

Maximizing Eq.~\eqref{opt_acq} represents a non-convex optimization problem over the domain $\mathcal{X}$. Since the constituent terms—including the GP posterior mean/variance and the weighted norm—are differentiable with respect to $x$ (assuming standard smooth kernels like Squared Exponential or Mat\'ern), the gradient of the acquisition function can be computed analytically or via automatic differentiation. To optimize this, we employ a multi-start gradient-based strategy using the L-BFGS algorithm, initialized with a diverse set of points sampled via a Latin Hypercube Design \citep{diessner2025nubo}. This approach helps mitigate the risk of converging to local optima inherent in the complex acquisition landscape. While our theoretical analysis assumes that this optimization oracle solves Eq.~\eqref{opt_acq} precisely without error, in practice, the solution is numerical and bounded by the optimizer's convergence tolerance. We assume this numerical error is negligible for the purpose of the regret analysis.

Once the optimal location $x_{t}$ is identified, we determine the appropriate fidelity level to query. The fidelity choice dictates the strategic intent: low-fidelity queries are utilized for exploration, while high-fidelity queries are reserved for exploitation. This decision relies on the confidence band of the low-fidelity model, defined as $CB(x,L) = \sqrt{\beta_{t}} \hat{\sigma}_{L,t}(x)$. A large value for $CB(x,L)$ implies high uncertainty regarding the low-fidelity function $f_{L}(x)$ at the chosen location. To reduce this uncertainty and improve the estimation for the high-fidelity, a low-fidelity query is preferred in such cases. Conversely, if the low-fidelity uncertainty is within an acceptable range, we query the high-fidelity source directly. Formally, if $CB(x,L)$ exceeds a fidelity selection threshold $\tau_{L}$—defined as the maximum tolerable epistemic uncertainty for the low-fidelity model—we query the low-fidelity source; otherwise, we proceed with the high-fidelity source to prevent over-sampling of the cheap fidelity. The fidelity acquisition function $m_{t}$ is defined as:
\begin{equation}\label{ACQ}
\begin{aligned}
m_{t}&=\begin{cases}
    L\,\,,\,\,CB(x,L)>\tau_{L}\\
    H\,\,,\,\,otherwise
\end{cases}
\end{aligned}
\end{equation}

For the final recommendation, we store
$\mathrm{FGP\text{-}LCB}_t(x)$ at each high-fidelity query.
At termination, provided $\mathcal{I}_H(\Lambda)\neq\varnothing$,
we return
\begin{equation}\label{eq:recommendation}
\hat{t}\in\operatorname{argmax}_{t\in\mathcal{I}_H(\Lambda)}
\mathrm{FGP\text{-}LCB}_t(x),
\qquad
x_{out}=x_{\hat{t}},
\end{equation}
with ties resolved by selecting the smallest index.
This rule selects the queried design with the largest lower confidence bound without requiring its latent high-fidelity value. The complete procedure is summarized in Algorithm 1 below.

\begin{algorithm}[h] 
   \caption{Fused Gaussian Process-based Upper Confidence Bound (FGP-UCB)}
\begin{algorithmic}
   \State{\bfseries Initialize:} $f_{L}\sim \mathcal{GP}(0,k_{L})$, $f_{\delta}\sim \mathcal{GP}(0,k_{\delta})$, $X_{L}$,$Y_{L}$,$X_{H}$,$Y_{H}$, $\mathcal{I}_H(\Lambda)$
   \For{$t=1$ to $T$}
       \State Choose a location $x_{t}=\argmax_{x}\;\text{FGP-UCB}_t(x)$ with $\text{FGP-UCB}_t(x)$ is defined in Eq.\eqref{UCB}
       \State Choose the fidelity $m_{t}$ based on the acquisition function in Eq.\eqref{ACQ}
       \State Observe outcome $y_{t}$ based on the chosen location $x_{t}$ and fidelity $m_t$
       \State $X_{m_t} = X_{m_t}\cup x_{t}$, $Y_{m_t} = Y_{m_t}\cup y_{t}$
       \If{$m_t = H$}
       \State Store $\text{FGP-LCB}_t(x_t)$, $\mathcal{I}_H(\Lambda) \gets \mathcal{I}_H(\Lambda) \cup \{t\}$
       \EndIf
       \State Update low-fidelity model $f_{L}$ with observations $X_{L}$, $Y_{L}$ in Eq.\eqref{muL}
       \State Update link function parameters $\rho$ and high-fidelity model $f_{H}$ with observations $X_{H}$, $Y_{H}$ and low-fidelity predictions in Eq.\eqref{eq3}
   \EndFor
   \State{\bfseries Output:} $x_{out}=x_{\hat{t}}$ according to Eq.~\eqref{eq:recommendation}
\end{algorithmic}
\end{algorithm}

\subsection{Regret Analysis}

In this section, we establish theoretical guarantees for the FGP-UCB algorithm by deriving bounds on the simple regret defined in Eq.\eqref{SR}.  While regret analysis is well-established for single-fidelity GP bandits~\citep{GPUCB}, the multi-fidelity setting introduces additional challenges in controlling fidelity allocation and the uncertainty of the fused model. The resulting bound depends on the maximum information gain of the residual GP over a high-value region of the design space.

Define the high-value region as
\begin{equation}
\mathcal{X}_{H} = \{x \in \mathcal{X} \mid f^{*} - f_{H}(x) \leq 3\tau_{L}\}
\end{equation}
where $\tau_L$ is the fidelity-selection threshold. Each point in $\mathcal X_H$ represents a near-optimal design: its latent high-fidelity objective value is close enough to $f^*$ that recommending it would incur simple regret of at most $3\tau_L$. A smaller $\tau_L$ requires designs in $\mathcal X_H$ to have latent high-fidelity objective values closer to $f^*$, which can shrink the high-value region.

Let $\Psi_n(\mathcal A)$ denote the maximum information gain from $n$ noisy observations of the residual GP at locations in $\mathcal A$, with noise variance $\sigma_H^2$. Because $\mathcal X_H\subseteq\mathcal X$, for the same residual kernel, noise variance, and number of observations,
\(
\Psi_n(\mathcal X_H)\leq\Psi_n(\mathcal X).
\)

When the confidence bounds hold, the upper confidence bound at each selected query location is at least $f^*$ because FGP-UCB maximizes that bound. The algorithm recommends the high-fidelity design with the largest stored lower confidence bound, so the value of the recommended design is at least as large as every stored lower bound. Thus, the width of the confidence interval at any high-fidelity query bounds the simple regret of the recommended design.

\begin{theorem}[Simple Regret Bound for FGP-UCB]\label{Theorem2}
Let $\delta\in(0,1)$ and $\mathcal{X}\subset [0,r]^{d}$ be compact and convex. There exists a finite $\Lambda_*$ such that, with probability at least $1-2\delta$, the design $x_{out}$ recommended by FGP-UCB satisfies
\begin{equation}\label{eq:localized_recommendation_regret}
S(\Lambda)=f^*-f_H(x_{out})
\leq
\sqrt{\frac{16K\beta_{\bar n_1}\Psi_{\bar n_H}(\mathcal X_H)}
{\bar n_H}},
\qquad\forall \Lambda\geq\Lambda_*,
\end{equation}
where $\bar n_1=\lfloor\Lambda/c_L\rfloor$, $\bar n_H=\lfloor\Lambda/c_H\rfloor$, $K>0$ is independent of $\Lambda$, and the exploration parameter $\beta_{t}$ is chosen according to the schedule specified in the Appendix.
\end{theorem}

Theorem~\ref{Theorem2} provides a high-probability bound on the simple regret of the design recommended by FGP-UCB under an evaluation budget $\Lambda$. The proof is provided in the Appendix. For a squared exponential residual kernel, $\Psi_n(\mathcal X_H)\leq\Psi_n(\mathcal X)=O((\log n)^{d+1})$~\citep{GPUCB}. With the logarithmic growth of $\beta_t$ under the specified schedule and fixed query costs, the regret bound therefore vanishes as $\Lambda$ increases. Beyond this convergence guarantee, the structure of our bound reveals several important insights when compared to the standard GP-UCB regret bound for single-fidelity optimization~\citep{GPUCB}:
\begin{enumerate}
    \item  \textbf{Reduced Effective Search Space}: The critical difference lies in the information gain term: $\Psi_{n}(\mathcal{X})$ for GP-UCB versus $\Psi_{\bar n_H}(\mathcal{X}_{H})$ for FGP-UCB. When the low-fidelity model is informative—meaning $f_L$ correlates well with $f_H$—the high-value region $\mathcal{X}_H$ constitutes a small fraction of the full domain. This translates directly to faster convergence.
    \item  \textbf{Volume Reduction}: When the correlation between fidelities is strong (i.e., the approximation $f_L \approx f_H$ is good), the high-value set $\mathcal{X}_{H}$ is significantly smaller than the full domain $\mathcal{X}$ (implying $\text{vol}(\mathcal{X}_{H}) \ll \text{vol}(\mathcal{X})$). Since $\Psi_T(\cdot)$ scales with the volume of the domain, this reduction leads to a tighter regret bound.
    \item  \textbf{Robustness}: In the worst-case scenario, where the lower fidelity provides little information, the set $\mathcal{X}_{H}$ expands to cover the majority of $\mathcal{X}$. In this limit, $\text{vol}(\mathcal{X}_{H}) \to \text{vol}(\mathcal{X})$, and the regret bound of FGP-UCB converges to that of standard GP-UCB. This guarantees that the multi-fidelity approach is robust: it offers accelerated convergence when correlations exist without performing worse than standard methods when they do not.
    \item \textbf{Novel Contribution}: To our knowledge, this is the \emph{first} simple regret bound for multi-fidelity Bayesian optimization using an adjustment-based surrogate model. An improvement over previous assumptions, which assumed independence among GPs with known correlation bounds~\citep{MFGPUCB} or used heuristic acquisition functions~\citep{poloczek2017multi, takeno2020multi}.
\end{enumerate}

\section{Experiments}\label{sec:Results}
This section evaluates the performance of the proposed algorithm using two multi-fidelity simulation studies, known as Six-Hump Camel and Park experiments, and a real-world case study involving experiment design for high-temperature superconductor (HTS) manufacturing. The query costs are set to $c_L = 1$ and $c_H = 10$ for the Six-Hump Camel and HTS experiments, and $c_L = 1$ and $c_H = 50$ for the Park experiments. In the simulation studies, the global high-fidelity optimum is known, and we therefore use the simple regret $S(\Lambda)$ which is a function of the query budget $\Lambda$ as the evaluation metric. Specifically, under each evaluation budget, the recommended design is selected from the high-fidelity queried points according to Eq. \eqref{eq:recommendation}, and its simple regret is evaluated using the underlying high-fidelity function without observation noise. In contrast, in the real-world experiment, the global optimum is not observable during deployment. Following the evaluation practice in applied multi-fidelity Bayesian optimization studies, we evaluate the observed high-fidelity response at the recommended design. 
This metric directly reflects the practical goal of HTS experiment design: identifying process conditions with high quality while reducing the number of expensive high-fidelity measurements.

\paragraph{Benchmark Models} To demonstrate the effectiveness of the proposed \textit{FGP-UCB} algorithm, we compare it against three representative baselines. The first baseline, \textit{GP-UCB}, performs standard single-fidelity Bayesian optimization by querying only high-fidelity samples~\citep{GP}. Although theoretically well-founded, it incurs significantly higher costs because it does not use inexpensive lower-fidelity information. The second baseline, \textit{MF-GP-UCB}, is a widely used multi-fidelity Gaussian Process bandit algorithm that models each fidelity independently using separate GPs; it does not incorporate the fused surrogate structure that underpins FGP-UCB~\citep{MFGPUCB}. The third baseline, \textit{MF-MES}, is an information-theoretic multi-fidelity Bayesian optimization approach that employs a multi-task GP surrogate and selects the next query by maximizing mutual information~\citep{takeno2020multi}. It is included as a representative heuristic acquisition-function method that does not offer theoretical performance guarantees. For a consistent comparison, all methods adopt the same final recommendation rule, which returns the high-fidelity queried point with the largest lower confidence bound.

\paragraph{Model Specification} All methods utilize a squared exponential kernel, $k(x,x^{'}) = \exp(-\frac{\Vert x-x^{'}\rVert^{2}}{2l^{2}})$ with zero-mean prior in Gaussian Process. The noise levels of observation used to generate simulated data are specified separately in Section \ref{sec:Simulation Study}. We initialize kernel hyperparameters by maximizing the GP marginal likelihood~\citep{williams2006gaussian} on the initial sample set and subsequently update them via marginal likelihood maximization. For MF-GP-UCB and FGP-UCB, the confidence band thresholds ($\tau_{L}$) are adaptively adjusted based on query ratios, as described in~\cite{MFGPUCB}. The parameter $\tau_{L}$ prevents excessive computational expenditure on low-fidelity queries. If $\tau_{L}$ is set too low, the algorithms may over-sample the low-fidelity source to reduce variance. We initialize $\tau_{L}$ with a small value; if the algorithm does not query the high-fidelity source for more than $c_{H}/c_{L}$ iterations, $\tau_{L}$ is doubled. The exploration-exploitation trade-off parameter ($\beta_{t}$) for GP-UCB, MF-GP-UCB, and FGP-UCB is set to $\beta_{t}=0.2d\log(2t)$~\citep{MFGPUCB}, where $d$ denotes the search space dimensionality. This setting accounts for scaling behavior with respect to both dimension and time.

\paragraph{Initialization} Following the recommendations in~\cite{brochu2010tutorial}, all models are initialized using a uniform random design. For the Six-Hump Camel and HTS experiments, where $c_L=1$ and $c_H=10$, FGP-UCB, MF-GP-UCB, and MF-MES are initialized with 3 high-fidelity and 20 low-fidelity observations, while GP-UCB begins with 5 high-fidelity observations, corresponding to an initialization cost of 50 for all methods. For the Park experiment, where $c_L=1$ and $c_H=50$, the multi-fidelity methods retain 3 high-fidelity and 20 low-fidelity observations, with an initialization cost of 170, while GP-UCB begins with 4 high-fidelity observations, with an initialization cost of 200. Each initialization is repeated 20 times, and the performance of each method is averaged over 20 independent runs to ensure statistical robustness. The shaded regions in the experimental figures represent the mean $\pm$ standard error of the mean (SEM) across 20 independent runs.

\subsection{Simulation Study} 
\label{sec:Simulation Study}
Two commonly used multi-fidelity simulation studies are considered:
\paragraph{Six-Hump Camel Function}
The Six-Hump Camel function~\citep{function} is a two-dimensional polynomial benchmark with an undulating surface of valleys separated by ridges, resembling a camel’s back. The surface contains six local minima, including two equally deep global minima located symmetrically on opposite sides of the origin. This study uses the true function surface as the latent high-fidelity function $f_H$ and constructs a correlated low-fidelity function $f_L$ by applying an input-dependent affine transformation to $f_H$. The input domains are $x_1 \in [-2,2]$ and $x_2 \in [-1,1]$. The functions are defined as:
\begin{equation}
\begin{aligned}
f_{H}(x)&= 4x_{1}^{2} - 2.1x_{1}^{4}+\frac{1}{3}x_{1}^{6}+x_{1}x_{2}-4x_{2}^{2}+4x_{2}^{4} \\
f_{L}(x)&= (0.5x_{1} - 0.3x_{2})f_{H} + (x_{1}-0.5)+2(x_{2}+0.5)+2
\end{aligned}
\end{equation}
\paragraph{Park Function}
The Park function is a four-dimensional analytical benchmark introduced by \cite{park1991tuning} for computer-code calibration and later adapted for multi-fidelity simulation studies. Its square-root, exponential, and trigonometric terms produce nonlinear interactions among the input variables. This four-dimensional model defines both high-fidelity and low-fidelity functions on the domain $[0,1]^{4}$. The functions are:
\begin{equation}
\begin{aligned}
f_{H}(x)=& \frac{x_{1}}{2}\sqrt{1+(x_{2}+x_{3}^{2})\frac{x_{4}}{x_{1}^{2}}}+(x_{1}+3x_{4}\exp(1+\sin(x_{3}))) \\
f_{L}(x)=& \left(1+\frac{\sin(x_{1})}{10}\right)f_{H}(x) - 2x_{1}^{2} + x_{2}^{2} + x_{3}^{2} + 0.5
\end{aligned}
\end{equation}

\begin{figure}
    \vspace{-2mm}
    \centering
    \begin{subfigure}[b]{0.49\textwidth}
        \centering
        \includegraphics[width=\linewidth]{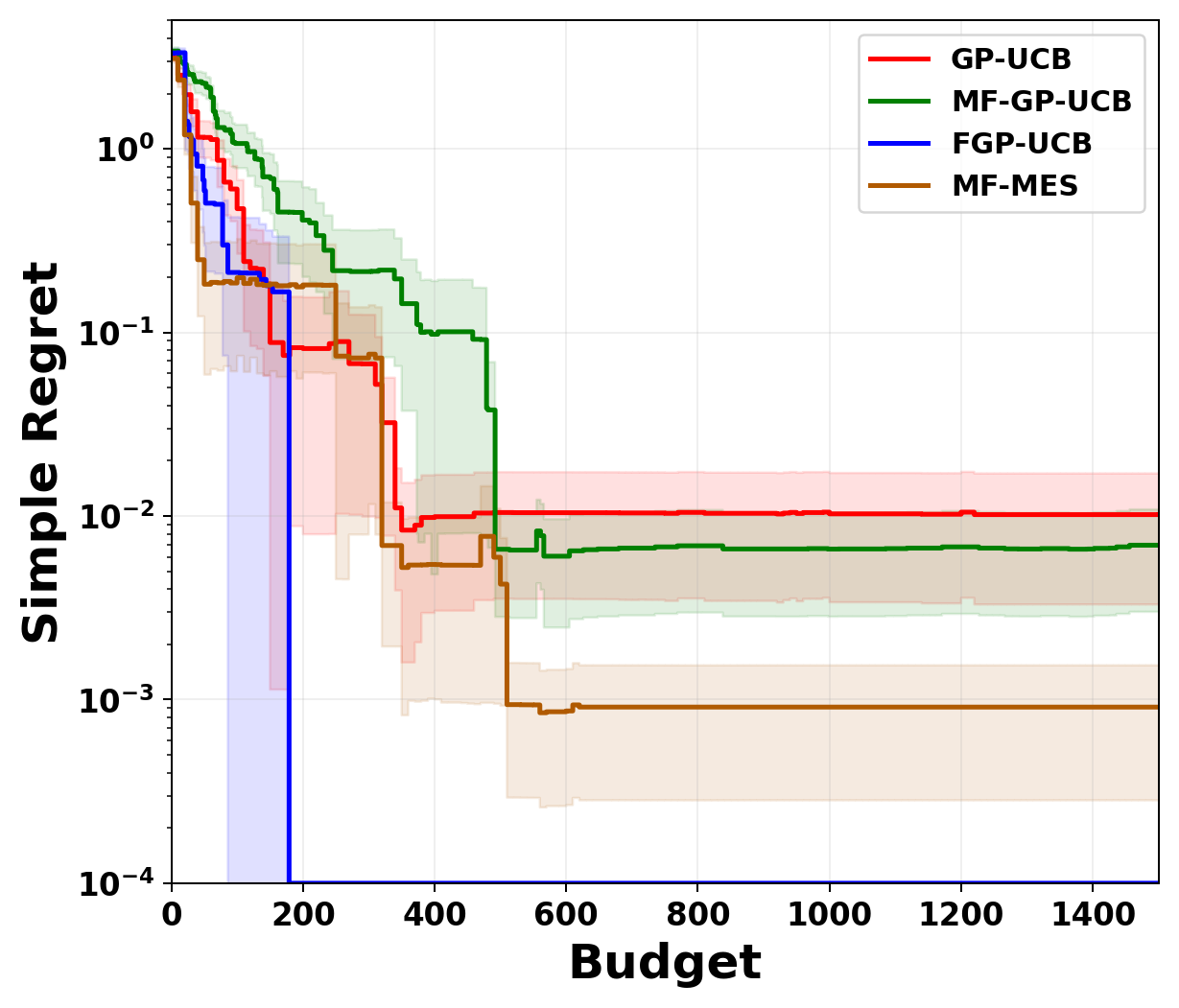}
        \caption{}
        \label{SR2D}
    \end{subfigure}
    \hfill 
    \begin{subfigure}[b]{0.49\textwidth}
        \centering
        \includegraphics[width=\linewidth]{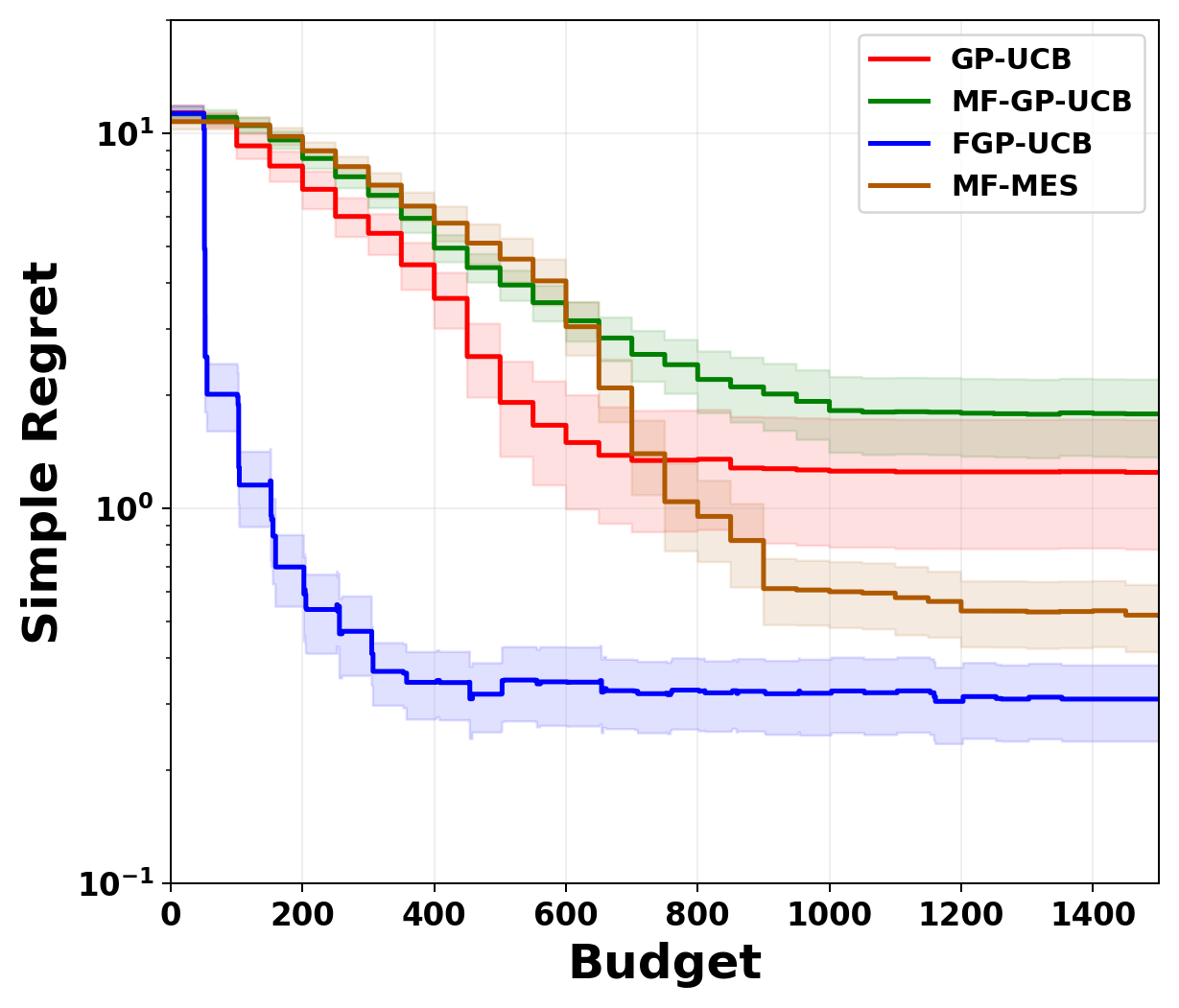}
        \caption{}
        \label{SR4D}
    \end{subfigure}    
    \caption{Simple regret of different methods under varying evaluation budgets in simulation studies. (a) Two-dimensional Six-Hump Camel function. (b) Four-dimensional Park function.}
    \label{fig:SR_Comparison}
    \vspace{-2mm}
\end{figure}

In both simulation models, noisy observations are generated independently at each fidelity as $y_m(x)=f_m(x)+\epsilon_m$, where $m\in{L,H}$, $\epsilon_L\sim\mathcal{N}(0,\sigma_L^2)$, and $\epsilon_H\sim\mathcal{N}(0,\sigma_H^2)$. The observation noise variances are set to $\sigma_L^2=0.1$ and $\sigma_H^2=0.01$, respectively. The total evaluation budget varies from 0 to 1500 after initialization, and the corresponding simple regret results across budgets are shown in Figure~\ref{fig:SR_Comparison}.

Figures \ref{SR2D} and \ref{SR4D} compare the performance of the different methods on the Six-Hump Camel simulation model and the Park function, respectively. Each curve reports the simple regret achieved under varying budget levels, averaged over 20 independent runs. In the Six-Hump Camel case, FGP-UCB rapidly reduces simple regret and reaches a near-zero level after a relatively small evaluation budget, indicating that the proposed method can quickly identify a near-optimal recommendation in a low-dimensional setting with informative low-fidelity structure. In the Park case, the optimization problem involves higher dimensionality, a more complex response surface, and a larger high-fidelity query cost, resulting in a more challenging optimization setting where the simple regret remains higher over the evaluated budget range. Nevertheless, FGP-UCB consistently achieves substantially lower simple regret than the baseline methods over most of the budget range. These results demonstrate the effectiveness of FGP-UCB in exploiting multi-fidelity information to identify high-quality recommendations under limited evaluation budgets.

\begin{figure}
    \vspace{-2mm}
    \centering
    \begin{subfigure}[b]{0.32\textwidth}
        \centering
        \includegraphics[width=\linewidth]{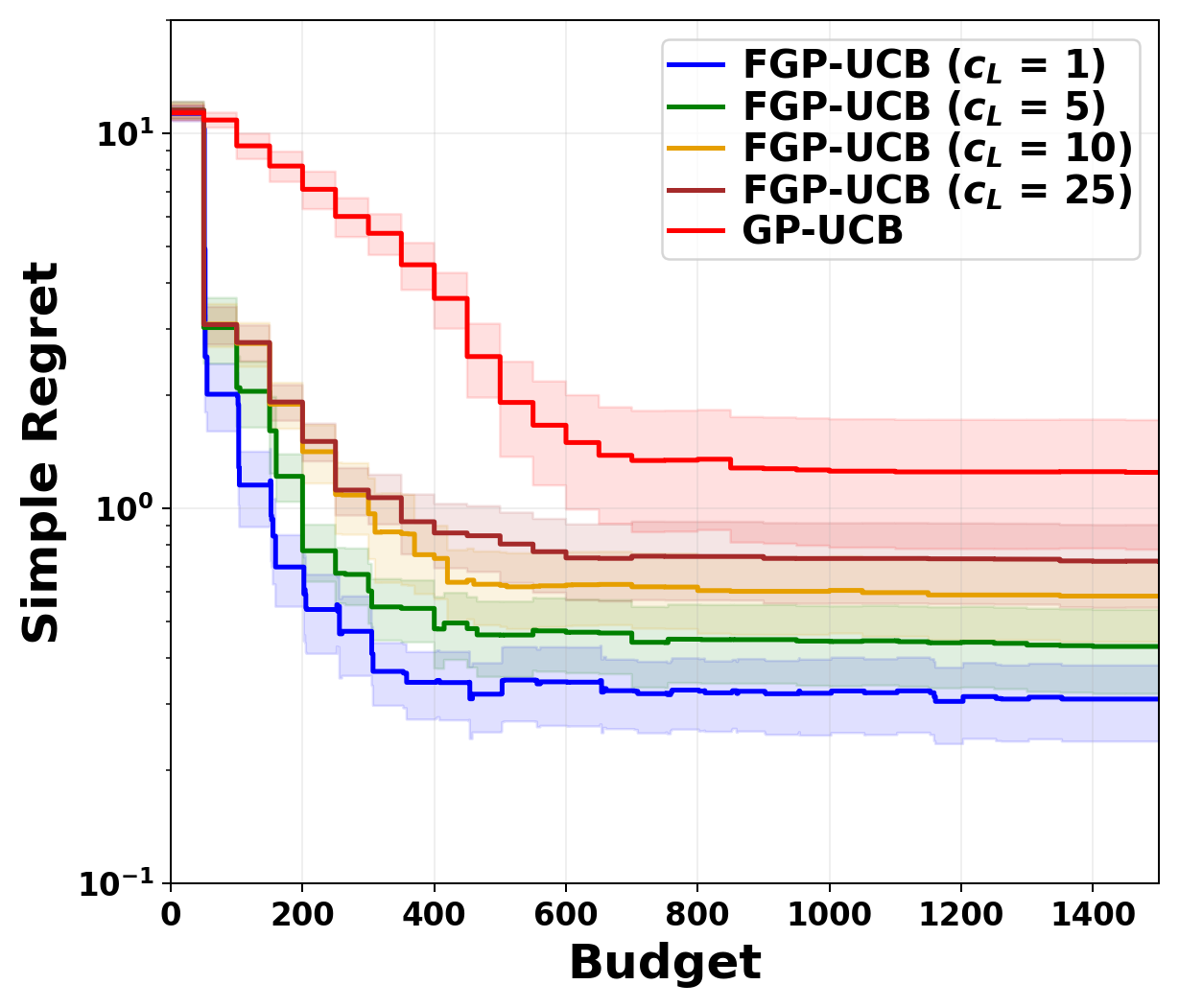}
        \caption{}
        \label{fig:cost}
    \end{subfigure}
    \hfill 
    \begin{subfigure}[b]{0.32\textwidth}
        \centering
        \includegraphics[width=\linewidth]{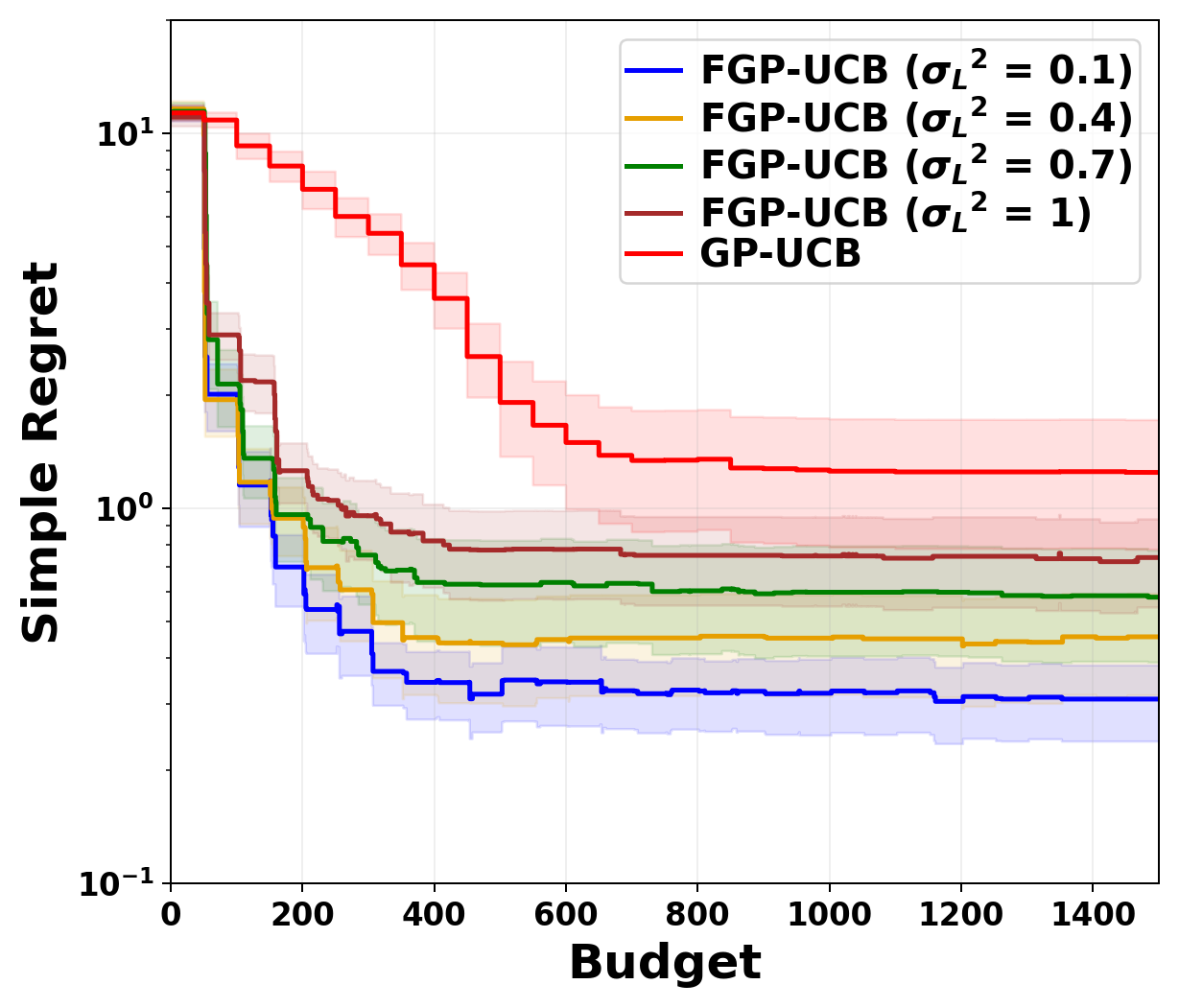}
        \caption{}
        \label{fig:noise}
    \end{subfigure}  
    \begin{subfigure}[b]{0.32\textwidth}
        \centering
        \includegraphics[width=\linewidth]{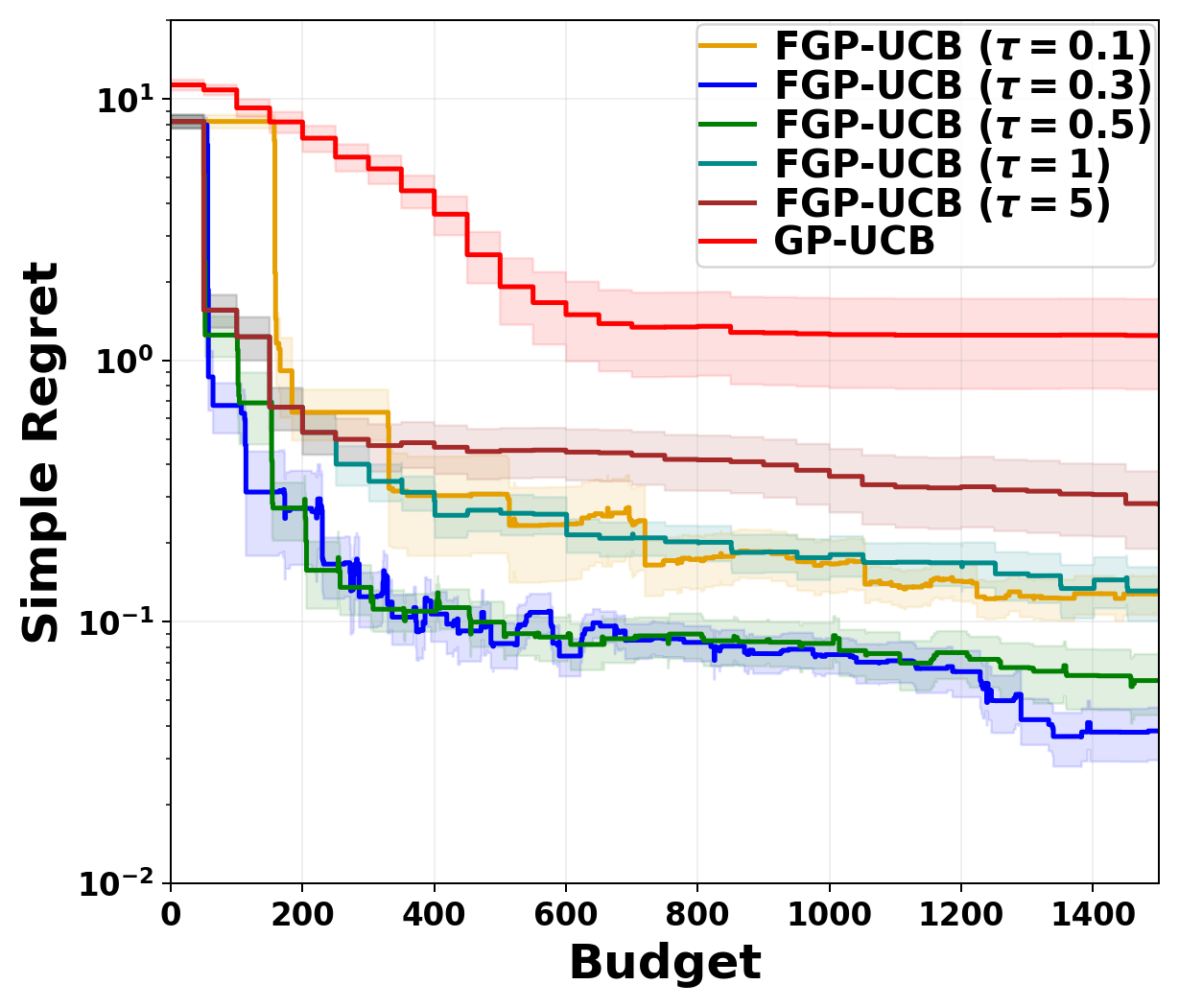}
        \caption{}
        \label{fig:threshold}
    \end{subfigure}  
    \caption{Sensitivity analysis of FGP-UCB on the four-dimensional Park function. (a) Effect of low-fidelity query cost ($c_L$). (b) Effect of low-fidelity observation noise variance ($\sigma_L^2$). (c) Effect of the fixed confidence band threshold ($\tau_L$).}
    \vspace{-2mm}
\end{figure}

\subsection{Sensitivity Analysis}
\paragraph{Effect of low-fidelity query cost} We first examine how the cost of low-fidelity evaluations influences the performance of the proposed FGP-UCB algorithm. Figure \ref{fig:cost} shows results on the four-dimensional, two-fidelity Park function for varying low-fidelity costs ($c_{L}$). The high-fidelity cost is fixed at $c_{H}=50$, while the low-fidelity cost varies from $c_{L}=[1,5,10,25]$. As $c_{L}$ increases, performance deteriorates. This decline arises because more expensive low-fidelity exploration consumes a larger portion of the budget, leaving fewer resources for high-fidelity evaluations and thereby slowing the reduction of simple regret. When the low-fidelity cost becomes relatively high ($c_L=25$), the performance of FGP-UCB is closest to GP-UCB, indicating that the benefit of exploiting low-fidelity information diminishes substantially as its cost increases.

\paragraph{Effect of low-fidelity noise} We next evaluate the algorithm’s sensitivity to the noise level in low-fidelity observations. Figure \ref{fig:noise} illustrates the performance of FGP-UCB across a range of low-fidelity noise variances $\sigma_{L}^{2}$ on the Park function. The high-fidelity noise variance is fixed at $\sigma_{H}^{2}=0.01$, while the low-fidelity noise variance ranges from $\sigma_{L}^{2}=0.1$ to $\sigma_{L}^{2}=1$. As $\sigma_L^2$ increases, low-fidelity observations become less informative, resulting in greater uncertainty in the low-fidelity predictions used by the fused surrogate. Consequently, the algorithm's performance deteriorates because less accurate low-fidelity estimates are less effective in guiding exploration, leading to slower convergence.

\paragraph{Effect of confidence band threshold}
We further examine how the confidence band threshold ($\tau_L$) influences the performance of FGP-UCB. Figure \ref{fig:threshold} compares the simple regret under different fixed thresholds on the Park function. Unlike the adaptive threshold strategy used in the previous experiments, $\tau_L$ is held constant throughout optimization to isolate its effect on fidelity selection. The threshold controls the balance between low-fidelity exploration and high-fidelity exploitation: a smaller threshold requires the low-fidelity uncertainty to be reduced further before a high-fidelity query is made, thereby encouraging more low-fidelity exploration, whereas a larger threshold allows the algorithm to switch to high-fidelity earlier and thus emphasizes exploitation. As shown in Figure \ref{fig:threshold}, intermediate threshold values ($\tau_L=0.3$ and $\tau_L=0.5$) achieve the best performance, while both smaller and larger thresholds lead to higher simple regret. These results indicate that an appropriate confidence band threshold is important for balancing the use of low- and high-fidelity evaluations and for achieving strong optimization performance.

\subsection{Real World Application}
High-temperature superconductor (HTS) tapes are pivotal to the development of next-generation high-field magnets due to their exceptional ability to carry high critical current. The critical current ($I_c$) of an HTS tape is not merely a material property—it is the primary quality metric, defining the maximum current the conductor can sustain without resistance. This parameter directly governs the performance, stability, and reliability of HTS tapes in real-world applications~\citep{SELVAMANICKAM}. However, determining how to tune process parameters—such as heating voltage, heating current, and related thermal conditions—to maximize $I_c$ remains a significant challenge in HTS manufacturing. In practice, $I_c$ is measured offline using a high-speed, high-resolution system that maps the magnetic field distribution along the tape: the reel-to-reel scanning Hall probe microscope (SHPM). While SHPM provides precise and spatially resolved measurements, it is time-consuming, expensive, and difficult to deploy frequently in a production environment, especially for long-length tapes. To alleviate these limitations, manufacturers often rely on in-line monitoring signals—such as substrate temperature—as indirect quality indicators~\citep{li2025quantile}. These measurements offer low-fidelity yet informative proxies for $I_c$. However, a key open question is how to systematically integrate these low-fidelity signals with sparse high-fidelity SHPM measurements to guide process control and optimization.

To address this challenge, we apply the proposed FGP-UCB method to fuse in-line monitoring data with offline quality measurements for optimal experiment design in HTS manufacturing. This approach provides a practical and cost-efficient strategy for identifying optimal experimental conditions while minimizing the need for costly high-fidelity evaluations.

\begin{figure}[t]
    \vspace{-2mm}
    \centering
    {\includegraphics[width=0.7\textwidth]{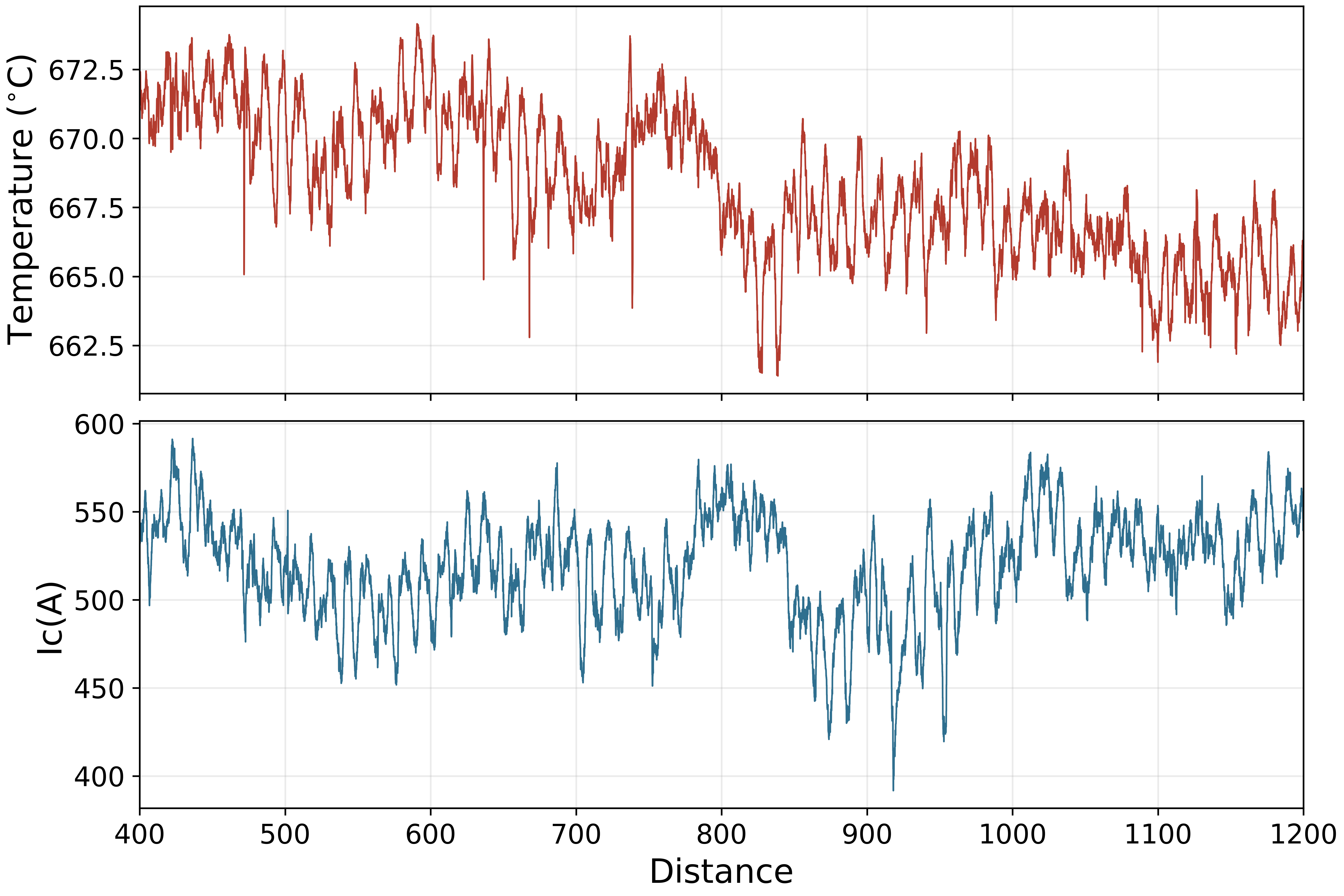}}
    \caption{Critical current ($I_c$) and Substrate temperature along the HTS tape length.}
    \vspace{-4mm}
    \label{M5vsiu}
\end{figure}

We evaluate FGP-UCB, GP-UCB, MF-GP-UCB, and MF-MES on a real-world HTS dataset containing various process settings—including heating voltage and current—along with their corresponding substrate temperature (low-fidelity) and $I_c$ (high-fidelity) measurements for a 0.8-meter HTS tape. Figure~\ref{M5vsiu} shows the profiles of $I_c$ and substrate temperature along the tape length. Both fluctuate due to variations in process parameters. The task is to identify the location achieving the maximum $I_c$ under a constrained evaluation budget. The costs of collecting substrate temperature and $I_c$ measurements are set to $c_{L}=1$ and $c_{H}=10$, respectively. Both signals are normalized prior to analysis, and initialization follows the same protocol used in the simulation study.

\subsubsection{Fused Gaussian Process Model Prediction}
Before evaluating the sequential optimization performance, we first examine whether the proposed fused Gaussian process surrogate can improve high-fidelity prediction when only a limited number of $I_c$ measurements are available. This analysis is motivated by the multi-fidelity prediction study in~\cite{tran2020multi}, where low-fidelity model outputs are fused with sparse high-fidelity observations to recover a more accurate high-fidelity response surface.

\begin{figure}[t]
    \centering
    \captionsetup[subfigure]{font=footnotesize}
    \begin{subfigure}{0.24\columnwidth}
        \centering
        \includegraphics[width=\linewidth]{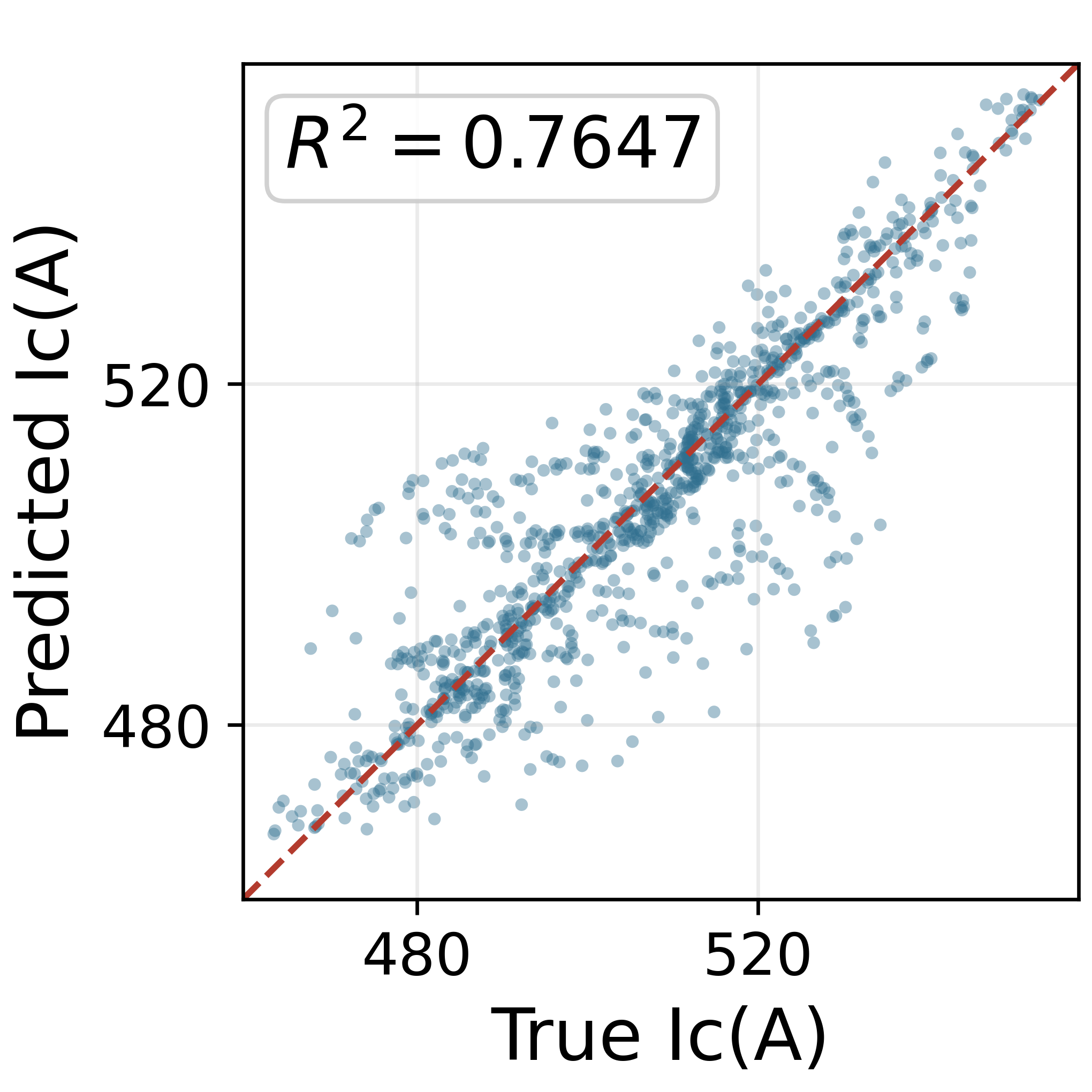}
        \caption{30 HF only}
        \label{fig:30HF}
    \end{subfigure}
    \hfill
    \begin{subfigure}{0.24\columnwidth}
        \centering
        \includegraphics[width=\linewidth]{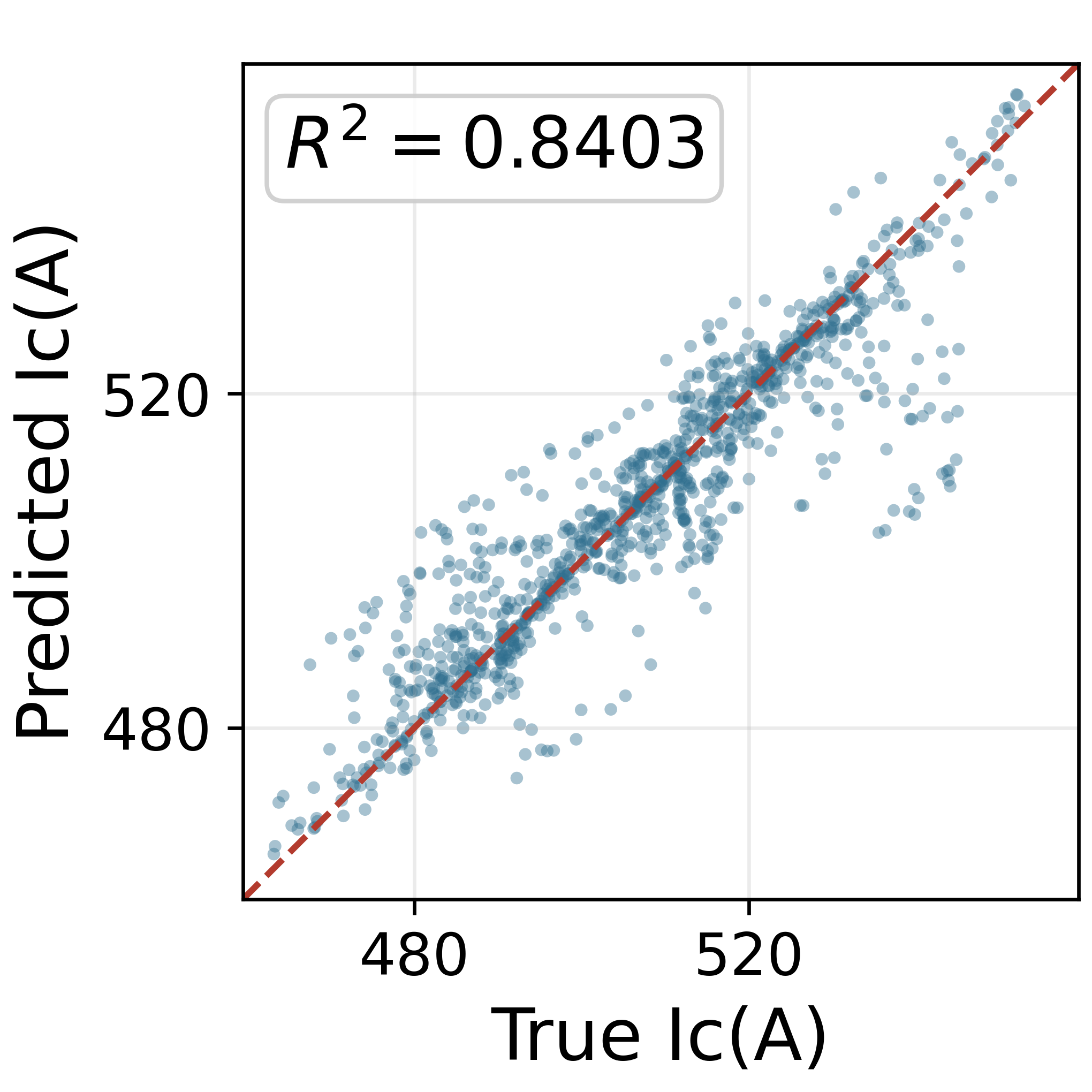}
        \caption{30 HF + 150 LF}
        \label{fig:30HF150LF}
    \end{subfigure}
    \hfill
    \begin{subfigure}{0.24\columnwidth}
        \centering
        \includegraphics[width=\linewidth]{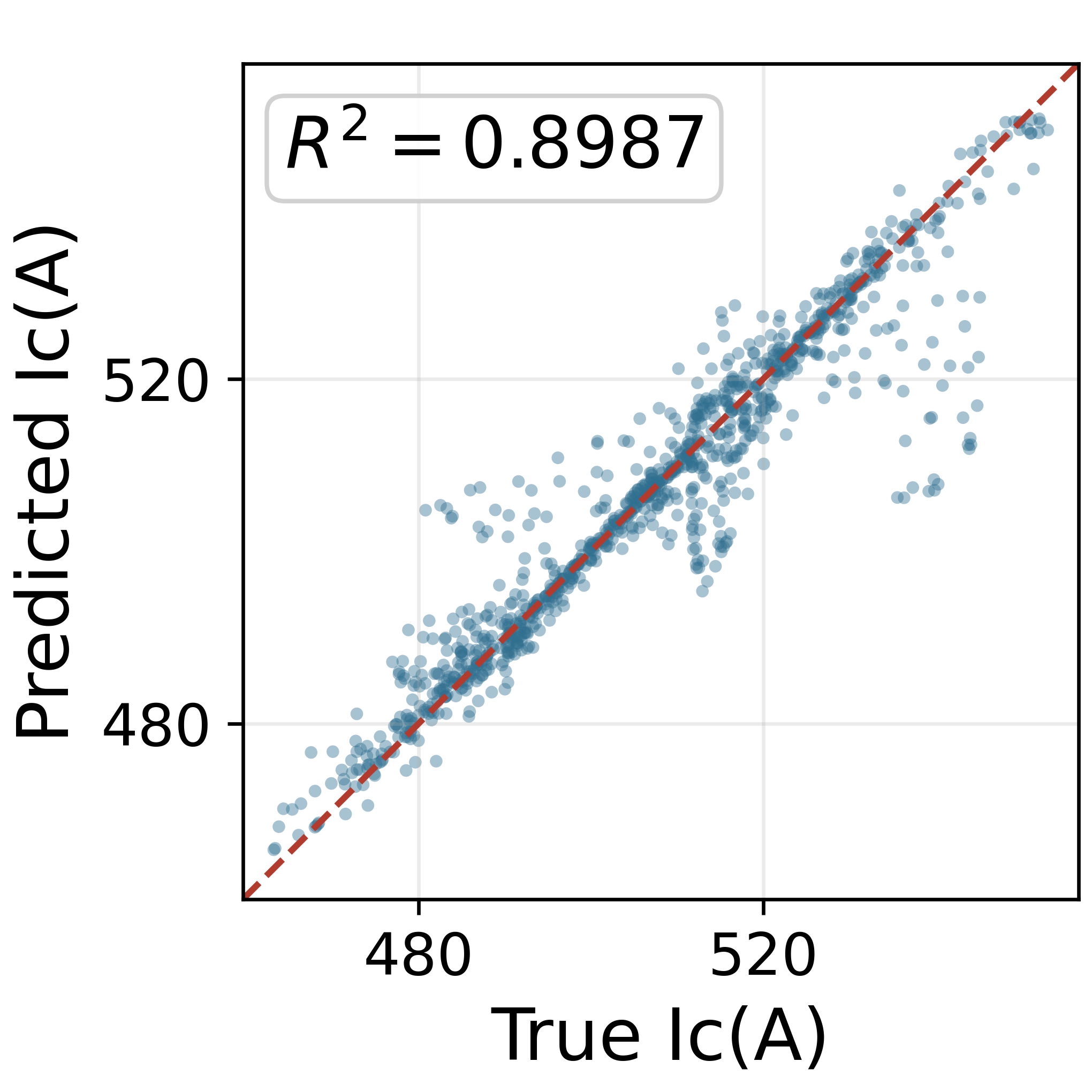}
        \caption{30 HF + 300 LF}
        \label{fig:30HF300LF}
    \end{subfigure}
    \hfill
    \begin{subfigure}{0.24\columnwidth}
        \centering
        \includegraphics[width=\linewidth]{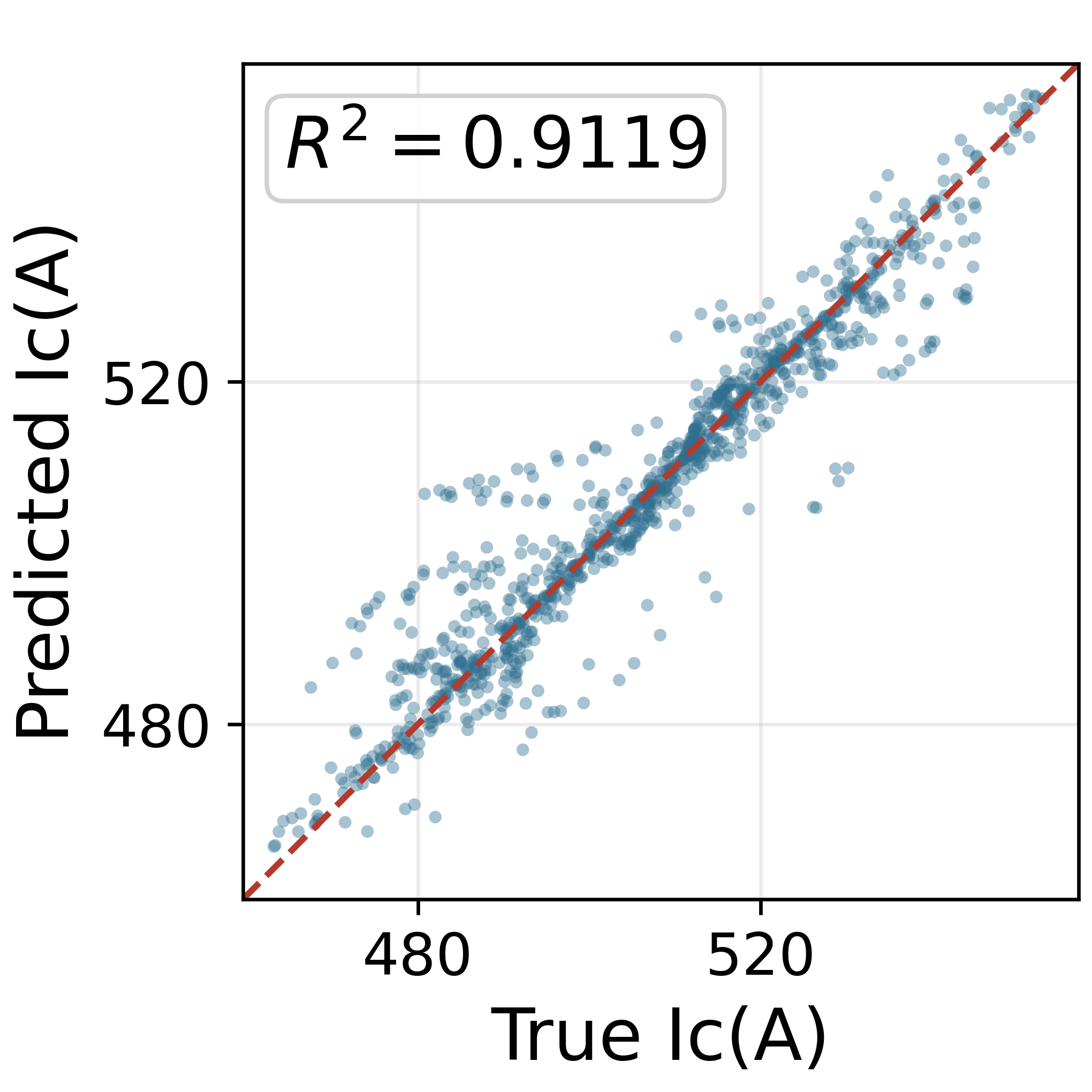}
        \caption{30 HF + 450 LF}
        \label{fig:30HF450LF}
    \end{subfigure}

    \caption{Comparison of High-fidelity model (\ref{fig:30HF}) and FGP multi-fidelity model (\ref{fig:30HF150LF}, \ref{fig:30HF300LF}, \ref{fig:30HF450LF}) prediction of High-temperature superconductor critical current $I_c$} 
    \label{fig:Prediction_compare}
\end{figure}

Figure~\ref{fig:Prediction_compare} compares predicted $I_c$ against observed high-fidelity $I_c$ under a fixed high-fidelity sample size. The first panel uses only 30 high-fidelity $I_c$ observations. The remaining panels keep the same number of high-fidelity observations but add increasing numbers of low-fidelity samples, from 150 to 300 and 450. The dashed diagonal represents perfect prediction. As the number of low-fidelity samples increases, the predicted points move closer to the diagonal and the $R^{2}$ value increases, indicating that the fused model more accurately recovers the high-fidelity $I_c$ surface.

These results show that, under scarce high-fidelity sampling, low-fidelity information provides useful structure for the FGP surrogate. The improvement comes from the fusion mechanism, which transfers information from the cheaper low-fidelity signal to improve prediction of the expensive high-fidelity response. This supports the use of the fused surrogate in the subsequent Bayesian optimization experiment.

\subsubsection{Optimization Performance}
In the HTS optimization experiment, we evaluate each method using the observed high-fidelity critical current ($I_c$) at its recommended design, since the true optimum is unknown in practice. Following the recommendation strategy in Section \ref{sec:FGPAQ}, the final design is selected from the high-fidelity queried points based on the maximum stored lower confidence bound. This evaluation protocol is consistent with applied multi-fidelity BO studies that assess optimization performance using observable high-fidelity responses rather than simple regret~\citep{mo2023accelerated}.

\begin{figure}
    \vspace{-2mm}
    \centering
    \begin{subfigure}[b]{0.49\textwidth}
        \centering
        \includegraphics[width=\linewidth]{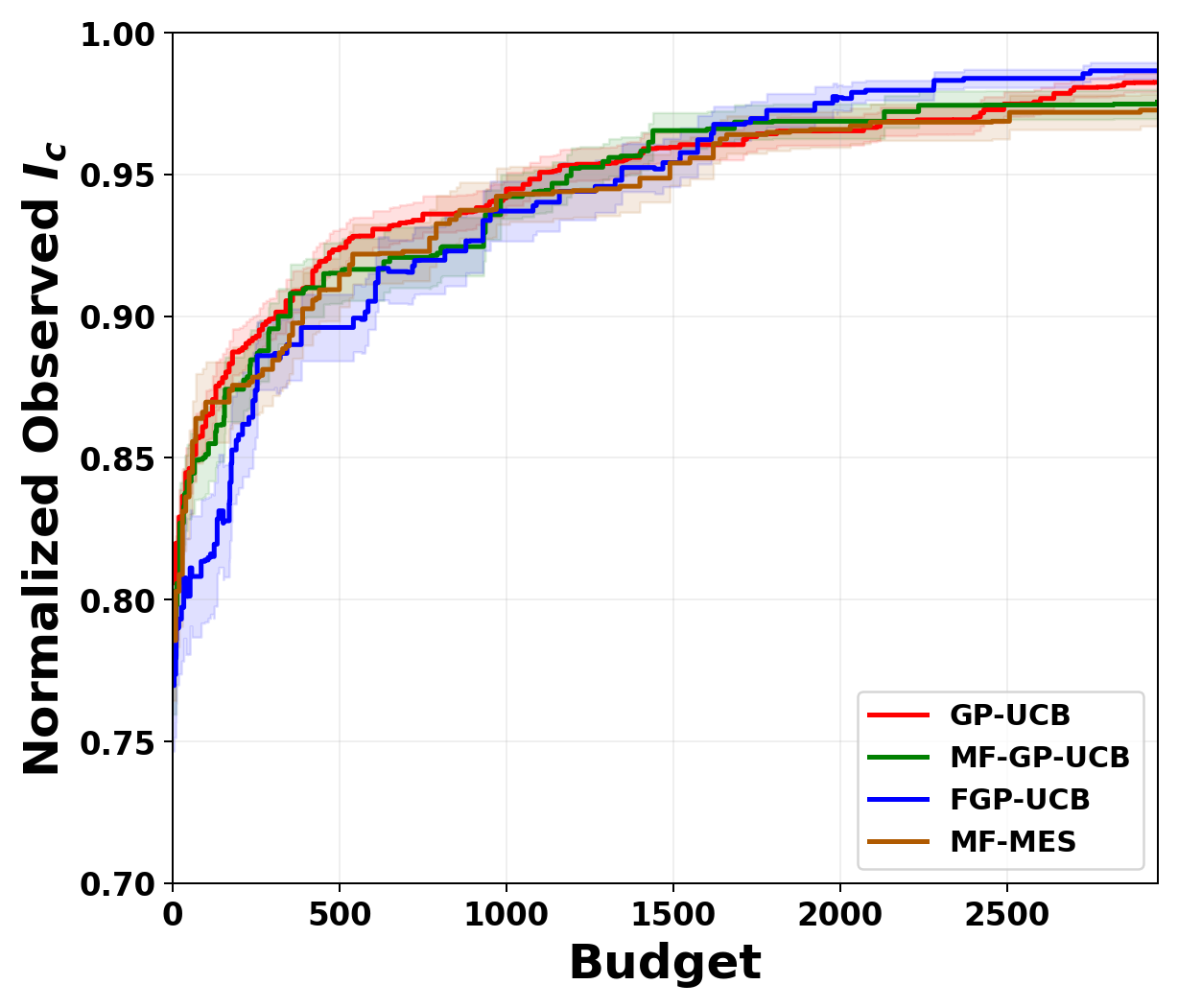}
        \caption{}
        \label{SRM5_full}
    \end{subfigure}
    \hfill 
    \begin{subfigure}[b]{0.49\textwidth}
        \centering
        \includegraphics[width=\linewidth]{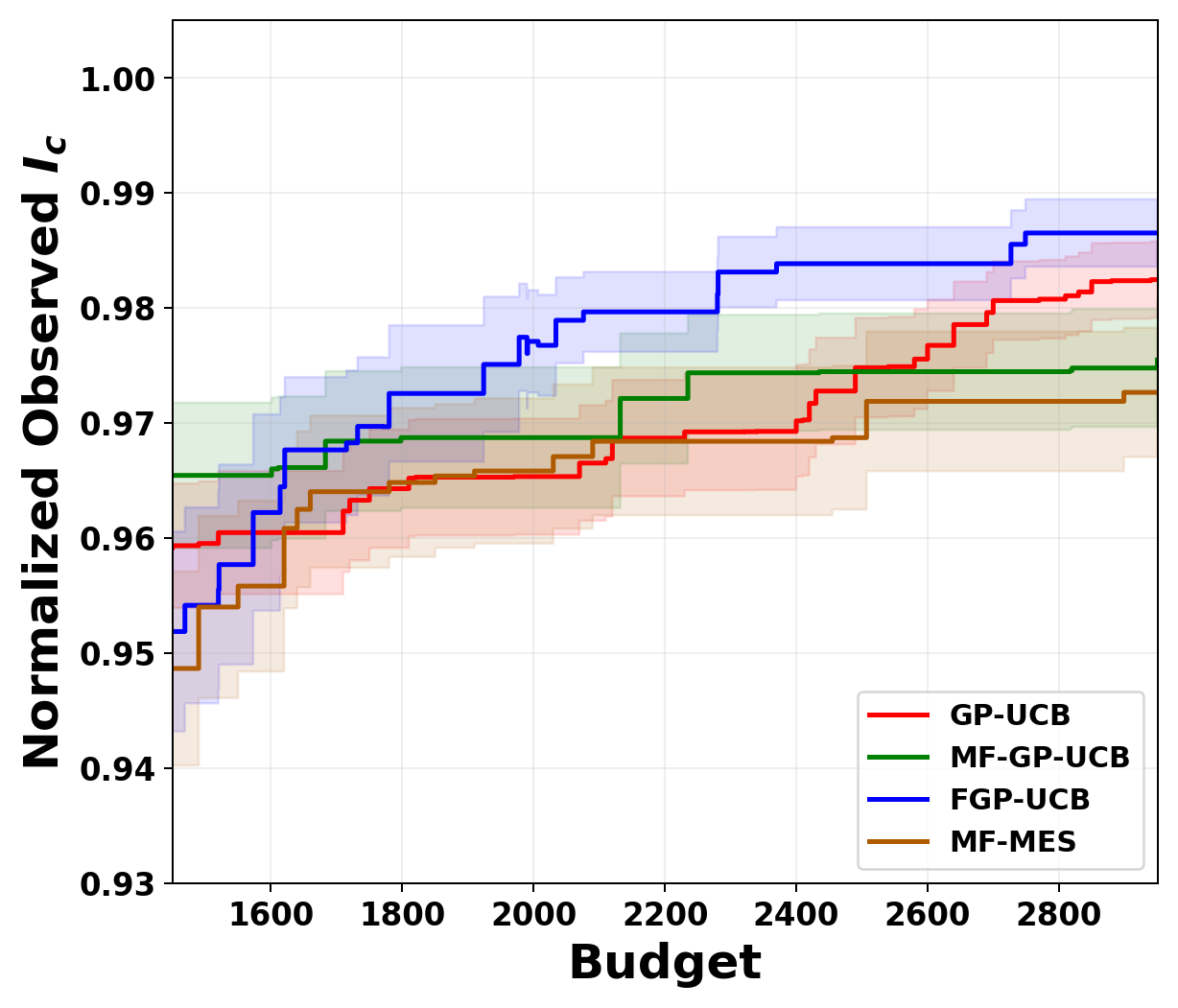}
        \caption{}
        \label{SRM5_zoom}
    \end{subfigure}    
    \caption{Observed high-fidelity critical current ($I_c$) at the recommended design under varying evaluation budgets for the HTS manufacturing experiment. (a) Performance over the full evaluation budget. (b) Enlarged view of the later stages of optimization.}
    \label{SRM5}
    \vspace{-2mm}
\end{figure}

Figure~\ref{SRM5} compares the optimization performance of the four methods in the HTS manufacturing application. Figure~\ref{SRM5_full} shows the results over the full evaluation budget, and Figure~\ref{SRM5_zoom} provides an enlarged view of the later stages. Each curve reports the observed high-fidelity critical current ($I_c$) at the recommended design as a function of the evaluation budget, averaged over 20 independent runs. GP-UCB leads early in the search. This early advantage may reflect how the methods allocate their budgets. GP-UCB uses its full budget to measure critical current directly, while the multi-fidelity methods can spend part of their early budgets on substrate temperature measurements that guide subsequent high-fidelity evaluations. FGP-UCB takes the lead later and achieves the highest average observed $I_c$ at the largest budget, followed by GP-UCB. The weaker performance of MF-GP-UCB may reflect its use of a uniform discrepancy bound, which can be overly conservative in regions where substrate temperature closely tracks critical current. MF-MES may spend part of its budget on less informative temperature measurements if its surrogate overestimates how much those measurements reveal about the maximum critical current. These results indicate that the proposed fused surrogate effectively leverages the correlation between substrate temperature and critical current, using inexpensive in-line monitoring information to guide the search toward high-quality regions while reducing reliance on expensive high-fidelity measurements.

\section{Conclusion}\label{sec:Conclusion}
This work addresses a fundamental gap in multi-fidelity Bayesian optimization: the lack of methods that simultaneously (i) exploit rich cross-fidelity correlations and (ii) provide rigorous regret guarantees. Existing decision-theoretic and surrogate-centric approaches—while effective in practice—do not optimize regret and consequently lack theoretical convergence assurances. Conversely, bandit-based algorithms such as MF-GP-UCB offer regret guarantees but rely on restrictive independence and bounded-difference assumptions that prevent them from fully leveraging informative low-fidelity data.

To bridge this gap, we introduced a novel algorithm, Fused Gaussian Process-based Upper Confidence Bound (FGP-UCB), that combines autoregressive multi-fidelity modeling with UCB-driven exploration, uncertainty-aware acquisition, and theoretically grounded regret performance. Theoretical analysis and simulations demonstrate that FGP-UCB effectively exploits lower-fidelity observations to inform the higher-fidelity model, resulting in improved regret performance compared to single-fidelity and existing MFBO methods. Beyond theoretical guarantees, we demonstrated the practical value of FGP-UCB by applying it to optimize process parameters in high-temperature superconductor (HTS) manufacturing, where it shows strong potential to enhance efficiency and reduce operational costs.

Future research could extend the proposed method to Bayesian optimization with multiple objectives and develop fused GP surrogates that capture dependencies across objectives and fidelity levels. Another direction is to accommodate low-fidelity approximation sources that evolve over time, such as self-improving LLM evaluators, by adapting the confidence bounds and fidelity-selection strategy to changes in their approximation accuracy.

\begin{ack}
We are grateful to University of Houston Advanced Manufacturing Institute for their dataset throughout the course of this work.
\end{ack}

\newpage

\bibliographystyle{plainnat}
\bibliography{ref}

\clearpage

\appendix

\section{Proof of Lemma~\ref{lemma1}}

The regularized plug-in log-likelihood is
\begin{equation}
\begin{aligned}
l_{\lambda}(\rho\mid\mathcal{D}_{H},\hat{\mu}_L,K_\delta)
={}&-\frac{1}{2}\log|Q|-\frac{\lambda}{2}\|\rho\|_2^2\\
&-\frac{1}{2}(Y_H-F_L\rho)^TQ^{-1}(Y_H-F_L\rho).
\end{aligned}
\end{equation}
With $Q=K_\delta+\sigma_H^2I$ and $A_L=F_L^TQ^{-1}F_L+\lambda I$,
maximization gives $\hat\rho=A_L^{-1}F_L^TQ^{-1}Y_H$. Under the
fusion model, $Y_H=F_L^*\rho^*+\eta_H$, where
$\eta_H=f_\delta(X_H)+\epsilon_H$. Writing
$b_\rho=A_L^{-1}F_L^TQ^{-1}(F_L^*-F_L)\rho^*$, we obtain
\[
\hat\rho-\rho^*-b_\rho
=A_L^{-1}\bigl(F_L^TQ^{-1}\eta_H-\lambda\rho^*\bigr).
\]
Since $A_L\succeq\lambda I$ and $\|\rho^*\|_2\leq M_\rho$,
the triangle inequality gives
\[
\|\hat\rho-\rho^*-b_\rho\|_{A_L}
\leq\|F_L^TQ^{-1}\eta_H\|_{A_L^{-1}}+\sqrt{\lambda}M_\rho.
\]
For fixed input locations and independent Gaussian observation noise,
$U=F_L^TQ^{-1}\eta_H$ is Gaussian with mean zero and covariance
$A_L-\lambda I$ conditional on the low-fidelity data. Consequently,
\[
\mathbb E\!\left[\exp\!\left(\tfrac12\|U\|_{A_L^{-1}}^2\right)
  \,\middle|\,\mathcal D_L\right]
=\det(A_L)^{1/2}\det(\lambda I)^{-1/2}.
\]
Markov's inequality gives the fixed-design concentration term
$\sqrt{2\log\bigl(\det(A_L)^{1/2}\det(\lambda I)^{-1/2}/\delta\bigr)}$
with unit scale. Establishing a radius for the adaptive design
requires a concentration argument for the refitted $F_{L,t}$.

\section{Proof of Theorem~\ref{theorem1}}

Let $b_{\rho,t}$ and $\eta_{H,t}$ denote the quantities above
computed from the data before round $t$. Write
\[
h_t(x)=x\hat\mu_{L,t}(x)-F_{L,t}^TQ_t^{-1}k_{\delta,t}(x),
\]
where $k_{\delta,t}(x)$ is the residual covariance vector.
Substituting $Y_{H,t}=F_{L,t}^*\rho^*+\eta_{H,t}$ into the fused
predictor gives
\[
\begin{aligned}
f_H(x)-\hat\mu_{H,t}(x)
={}&h_t(x)^T(\rho^*-\hat\rho_t+b_{\rho,t})\\
&+((\rho^*)^Tx)\bigl(f_L(x)-\hat\mu_{L,t}(x)\bigr)\\
&-k_{\delta,t}(x)^TQ_t^{-1}(F_{L,t}^*-F_{L,t})\rho^*
  -h_t(x)^Tb_{\rho,t}\\
&+f_\delta(x)-k_{\delta,t}(x)^TQ_t^{-1}\eta_{H,t}.
\end{aligned}
\]
Cauchy--Schwarz bounds the absolute value of the first term by
$\|h_t(x)\|_{A_{L,t}^{-1}}\|\hat\rho_t-\rho^*-b_{\rho,t}\|_{A_{L,t}}$.
The residual GP confidence event requires the two-sided bound
\[
\bigl|f_\delta(x)-k_{\delta,t}(x)^TQ_t^{-1}\eta_{H,t}\bigr|
\leq\sqrt{\beta_t}\,\hat\sigma_{\delta,t}(x)
\]
using $\eta_{H,t}$ as the residual observations. Establishing the
radius in Theorem~\ref{theorem1} also requires controlling the
low-fidelity substitution terms and the link contribution
through $h_t(x)$.

\section{Proof of Theorem~\ref{Theorem2}}\label{app:proof_theorem2}

\begin{lemma}\label{lemma C6}
Let $\{z_s\}_{s=1}^m\subset\mathbb R^d$ satisfy
$\|z_s\|_2\leq G$, and let
$V_s=\lambda I+\sum_{i=1}^s z_i z_i^T$, where $\lambda>0$.
Then
\[
\det(V_m)\leq\left(\lambda+\frac{mG^2}{d}\right)^d,
\qquad
\sum_{s=1}^m\|z_s\|_{V_{s-1}^{-1}}^2
\leq\left(1+\frac{G^2}{\lambda}\right)
d\log\left(1+\frac{mG^2}{\lambda d}\right).
\]
In particular, the factor $1+G^2/\lambda$ can be replaced by $2$
when $\lambda\geq G^2$.
\end{lemma}

\begin{proof}
The determinant bound follows from
$\operatorname{tr}(V_m)\leq\lambda d+mG^2$ and the
arithmetic--geometric mean inequality. Write
$u_s=z_s^TV_{s-1}^{-1}z_s\leq G^2/\lambda$.
Since $u\leq(1+G^2/\lambda)\log(1+u)$ for
$0\leq u\leq G^2/\lambda$, the matrix determinant lemma gives
\[
\sum_{s=1}^m u_s
\leq\left(1+\frac{G^2}{\lambda}\right)
\log\frac{\det(V_m)}{\det(\lambda I)}
\leq\left(1+\frac{G^2}{\lambda}\right)
d\log\left(1+\frac{mG^2}{\lambda d}\right).
\]
\end{proof}

\begin{lemma}\label{informationgain}
For a fixed sequence $Z=(x_1,\ldots,x_m)$ and a GP with kernel
$k$ and independent Gaussian observation noise of variance
$\sigma^2>0$, let $s_{i-1}^2(x_i)$
be the predictive variance based only on $x_1,\ldots,x_{i-1}$. Then
\[
I(y_Z;f_Z)
=\frac12\log\det(I_m+\sigma^{-2}K_k(Z,Z))
=\frac12\sum_{i=1}^m\log(1+\sigma^{-2}s_{i-1}^2(x_i)).
\]
If $k(x,x)\leq1$, then for any subset $\mathcal J$ of iterations
that query this GP with $x_t\in\mathcal A$ and $|\mathcal J|=m$,
\begin{equation}\label{eq:appendix_subset_variance}
\sum_{t\in\mathcal J}\hat\sigma_t^2(x_t)
\leq\frac{2}{\log(1+\sigma^{-2})}
\Psi_m^{k,\sigma}(\mathcal A),
\end{equation}
where $\hat\sigma_t^2$ uses all available observations of that GP
before iteration $t$, and $\Psi_m^{k,\sigma}(\mathcal A)$ is the
supremum of the information gain over sequences of $m$ points in
$\mathcal A$, allowing repeated locations.
\end{lemma}

\begin{proof}
The determinant identity follows by successive Schur complements.
Order the points in $\mathcal J$ chronologically. Conditioning on
additional observations can only decrease the predictive variance, so
$\hat\sigma_t^2(x_t)\leq s_{i-1}^2(x_i)\leq1$ for the corresponding
point. The inequality
\[
v\leq\frac{\log(1+\sigma^{-2}v)}{\log(1+\sigma^{-2})},
\qquad 0\leq v\leq1,
\]
and the definition of MIG yield~\eqref{eq:appendix_subset_variance}.
The determinant and variance inequalities apply to every realized
sequence, including adaptively selected points and subsets.
\end{proof}

\paragraph{Assumption 1 (Stable link information).}
Let $t_j$ denote the round of the $j$th high-fidelity query after
initialization and let $z_j=x_{t_j}\hat\mu_{L,t_j}(x_{t_j})$
be its feature computed before the query. For some constant
$\kappa>0$ independent of $\Lambda$, assume
\[
A_{L,t_j}\succeq\kappa\left(\lambda I+\sum_{i=1}^{j-1}z_i z_i^T\right)
\qquad\text{for all }j.
\]

\paragraph{Additional conditions.}
The kernels and noise variances are fixed, with
$k_L(x,x)\leq1$, $k_\delta(x,x)=1$, and
$\sigma_L^2,\sigma_H^2>0$. The predictive variances are computed
using the corresponding GP covariance formulas. The constants
$\lambda,\tau_L$ are positive, $0<c_L<c_H$, and $\beta_t$ is
positive and nondecreasing. For constants
$G,a_\rho>0$ independent of $\Lambda$,
\begin{equation}\label{eq:appendix_link_bounds}
\|z_j\|_2\leq G,
\qquad
\alpha_{\rho,t_j}^2\leq a_\rho\beta_{t_j}
\qquad\text{for all }j.
\end{equation}
Assume that Assumption~1 and~\eqref{eq:appendix_link_bounds} hold
jointly on an event $\mathcal E_{\rm link}$ of probability at least
$1-\delta$. Let $\mathcal E_{\rm conf}$ be the event that the
two-sided bound in Theorem~\ref{theorem1} holds simultaneously for
all $t$ and $x\in\mathcal X$. With simultaneous confidence at level
$1-\delta$, the union bound gives
$\Pr(\mathcal E)\geq1-2\delta$ for
$\mathcal E=\mathcal E_{\rm link}\cap\mathcal E_{\rm conf}$.
Write $\Psi_n^L=\Psi_n^{k_L,\sigma_L}$ and
$\Psi_n=\Psi_n^{k_\delta,\sigma_H}$. Also assume, as $\Lambda\to\infty$,
\begin{equation}\label{eq:appendix_mig_growth}
\beta_{\bar n_1}\Psi_{\bar n_1}^L(\mathcal X)=o(\bar n_1),
\qquad
\beta_{\bar n_1}\Psi_{\bar n_H}(\mathcal X)=o(\bar n_H),
\end{equation}
where $\bar n_1=\lfloor\Lambda/c_L\rfloor$ and
$\bar n_H=\lfloor\Lambda/c_H\rfloor$.
The algorithm maximizes the FGP-UCB score exactly and continues
until the remaining budget cannot pay for its next query. A fixed
initialization cost, if any, is denoted by $\Lambda_0$.

\begin{proof}[Proof of Theorem~\ref{Theorem2}]
Work on $\mathcal E$. Let $N_L$ and $N_H$ denote the numbers of
low- and high-fidelity queries after initialization, and let
$\mathcal I_H(\Lambda)$ be the corresponding high-fidelity query
indices. Then $T=N_L+N_H\leq \bar n_1$ and $N_H\leq \bar n_H$.
All predictions below are computed before the indexed query.
Write the confidence radius as
\[
w_t(x)=\alpha_{\rho,t}
\|x\hat\mu_{L,t}(x)\|_{A_{L,t}^{-1}}
+\sqrt{\beta_t}\,\hat\sigma_{\delta,t}(x).
\]
The confidence bound and FGP-UCB maximization imply
\begin{equation}\label{eq:appendix_query_gap}
f^*\leq\mathrm{FGP\text{-}UCB}_t(x_t)
=\hat\mu_{H,t}(x_t)+w_t(x_t),
\qquad
f^*-f_H(x_t)\leq2w_t(x_t).
\end{equation}

\paragraph{Accumulated uncertainty.}
Consider any nonempty subset
$\mathcal J\subseteq\mathcal I_H(\Lambda)$ of size $m$, with all
its points in a nonempty set $\mathcal A$. Write its ordered
high-fidelity indices as $t_{j_1},\ldots,t_{j_m}$ and define
$W_\ell=\lambda I+\sum_{h=1}^{\ell}z_{j_h}z_{j_h}^T$.
Because $\{j_1,\ldots,j_{\ell-1}\}\subseteq\{1,\ldots,j_\ell-1\}$,
Assumption~1 gives
\[
A_{L,t_{j_\ell}}\succeq\kappa W_{\ell-1}.
\]
Hence Lemma~\ref{lemma C6} and~\eqref{eq:appendix_link_bounds}
yield
\begin{equation}\label{eq:appendix_link_sum}
\sum_{\ell=1}^m\alpha_{\rho,t_{j_\ell}}^2
\|z_{j_\ell}\|_{A_{L,t_{j_\ell}}^{-1}}^2
\leq\frac{a_\rho\beta_{\bar n_1}d}{\kappa}
\left(1+\frac{G^2}{\lambda}\right)
\log\left(1+\frac{mG^2}{\lambda d}\right).
\end{equation}
Repeated sampling at any point of $\mathcal A$ gives
$\Psi_m(\mathcal A)\geq\tfrac12\log(1+m/\sigma_H^2)$.
Using $\log(1+av)\leq\max\{1,a\}\log(1+v)$ for $a,v\geq0$,
define
\[
C_\rho=\frac{2a_\rho d}{\kappa}
\left(1+\frac{G^2}{\lambda}\right)
\max\left\{1,\frac{\sigma_H^2G^2}{\lambda d}\right\},
\qquad
C_\delta=\frac{2}{\log(1+\sigma_H^{-2})},
\qquad
K=(\sqrt{C_\rho}+\sqrt{C_\delta})^2.
\]
Thus~\eqref{eq:appendix_link_sum} is at most
$C_\rho\beta_{\bar n_1}\Psi_m(\mathcal A)$, while
Lemma~\ref{informationgain} gives
\[
\sum_{t\in\mathcal J}\beta_t\hat\sigma_{\delta,t}^2(x_t)
\leq C_\delta\beta_{\bar n_1}\Psi_m(\mathcal A).
\]
The triangle inequality for the Euclidean norm therefore implies
\begin{equation}\label{eq:appendix_width_sum}
\sum_{t\in\mathcal J}w_t(x_t)^2
\leq K\beta_{\bar n_1}\Psi_m(\mathcal A).
\end{equation}

\paragraph{Number of high-fidelity queries in $\mathcal X_H$.}
A low-fidelity query requires
$\sqrt{\beta_t}\hat\sigma_{L,t}(x_t)>\tau_L$.
With $C_L=2/\log(1+\sigma_L^{-2})$,
Lemma~\ref{informationgain} yields
\[
N_L\tau_L^2
\leq\beta_{\bar n_1}\sum_{t:\,m_t=L}\hat\sigma_{L,t}^2(x_t)
\leq C_L\beta_{\bar n_1}\Psi_{\bar n_1}^L(\mathcal X).
\]
Define
\[
B_L(\Lambda)=\frac{C_L\beta_{\bar n_1}\Psi_{\bar n_1}^L(\mathcal X)}{\tau_L^2}.
\]
The unspent budget is less than $c_H$, so
\[
c_HN_H>\Lambda-\Lambda_0-c_H-c_LN_L
\geq\Lambda-\Lambda_0-c_H-c_LB_L(\Lambda).
\]
Since $B_L(\Lambda)=o(\Lambda)$, for all sufficiently large
$\Lambda$ this gives $N_H\geq \bar n_H/2$.

Let $N_{\rm out}$ count the high-fidelity queries outside
$\mathcal X_H$. For each such query,
$3\tau_L<f^*-f_H(x_t)\leq2w_t(x_t)$.
By~\eqref{eq:appendix_width_sum} with $\mathcal A=\mathcal X$,
\[
N_{\rm out}
\leq\frac{4}{9\tau_L^2}\sum_{t\in\mathcal I_H(\Lambda)}w_t(x_t)^2
\leq\frac{4K\beta_{\bar n_1}\Psi_{\bar n_H}(\mathcal X)}{9\tau_L^2}
=o(\bar n_H).
\]
Consequently, there is a finite $\Lambda_*$ such that, for every
$\Lambda\geq\Lambda_*$,
\begin{equation}\label{eq:appendix_inside_count}
N_{\rm in}:=N_H-N_{\rm out}\geq\frac{\bar n_H}{4}>0.
\end{equation}

\paragraph{Regret of the recommendation.}
Let $\mathcal J_H=\{t\in\mathcal I_H(\Lambda):x_t\in\mathcal X_H\}$,
so $|\mathcal J_H|=N_{\rm in}$. The stored FGP-LCB recommendation
and the confidence bound give, for every $t\in\mathcal J_H$,
\[
\begin{aligned}
f_H(x_{out})
&\geq\max_{s\in\mathcal I_H(\Lambda)}
\mathrm{FGP\text{-}LCB}_s(x_s)\\
&\geq\mathrm{FGP\text{-}LCB}_t(x_t)
=\hat\mu_{H,t}(x_t)-w_t(x_t).
\end{aligned}
\]
Combining this with~\eqref{eq:appendix_query_gap} gives
$S(\Lambda)\leq2w_t(x_t)$ for every $t\in\mathcal J_H$. Therefore,
by~\eqref{eq:appendix_width_sum} and~\eqref{eq:appendix_inside_count},
\[
\begin{aligned}
S(\Lambda)
&\leq2\min_{t\in\mathcal J_H}w_t(x_t)
\leq2\sqrt{\frac{\sum_{t\in\mathcal J_H}w_t(x_t)^2}{N_{\rm in}}}\\
&\leq\sqrt{\frac{4K\beta_{\bar n_1}\Psi_{N_{\rm in}}(\mathcal X_H)}{N_{\rm in}}}
\leq\sqrt{\frac{16K\beta_{\bar n_1}\Psi_{\bar n_H}(\mathcal X_H)}{\bar n_H}}.
\end{aligned}
\]
Finally, $\Psi_{\bar n_H}(\mathcal X_H)\leq\Psi_{\bar n_H}(\mathcal X)$
and~\eqref{eq:appendix_mig_growth} imply $S(\Lambda)\to0$.
All inequalities hold on $\mathcal E$.
\end{proof}

\clearpage


\end{document}